\documentclass[runningheads]{llncs}

\usepackage[T1]{fontenc}
\usepackage{amsmath,amssymb,mathtools}
\usepackage{mathrsfs}
\usepackage{graphicx}
\usepackage{booktabs}
\usepackage{tabularx}
\usepackage{array}
\usepackage{float}
\usepackage{multirow}
\usepackage{makecell}
\usepackage{adjustbox}
\usepackage{threeparttable}
\usepackage{algorithm}
\usepackage{algorithmic}
\usepackage[hidelinks]{hyperref}

\title{Alignment Flywheel-Guided Suppression Patching for Interactionless Inverse Reinforcement Learning}
\titlerunning{Flywheel-Guided Suppression Patching}

\author{Elias Malomgr\'e \and Pieter Simoens}
\authorrunning{E. Malomgr\'e and P. Simoens}
\institute{IDLab, Ghent University -- imec, Ghent, Belgium\\
\email{\{elias.malomgre,pieter.simoens\}@ugent.be}}

\begin{document}

\maketitle
\begin{abstract}
Learned reward models allow reinforcement-learning agents to use demonstrations,
implicit feedback, safety constraints, social norms, or other sources that do
not provide a complete hand-designed scalar reward. However, a learned reward model can assign high reward to unwanted
behavior. A policy can exploit these errors during training. Later
requirements can also identify additional regions where the deployed reward
should change. We introduce \emph{flywheel-guided patching}, a mechanism for
making such localized corrections without retraining the learned evaluator. An
Alignment Flywheel identifies overgeneralized or constraint-violating regions
and compiles verified findings into suppression patches. We instantiate the
mechanism for Interactionless Inverse Reinforcement Learning using MoE-GUIDE, placing
patches in the model's latent space and retrieving only a small local subset
during online reward evaluation. In the main MuJoCo benchmark, we evaluate patching without
expert-state reset warm starts. We also test whether operator-specified norms absent
from the demonstrations can be compiled into the same reward interface and
compare this with enforcing the corresponding rules online. The results show
that localized corrective information can modify an existing learned reward
without providing a complete replacement reward. Compilation can reuse judgments that are expensive or unavailable
during control. Cheap deterministic rules may be simpler to check directly
at each step.
Code for MoE-GUIDE, Alignment Flywheel, and flywheel patching is at https://github.com/decide-ugent/Alignment-Flywheel.
\end{abstract}

\section{Introduction}
\label{sec:intro}

One of the main problems in reinforcement learning (RL) \cite{sutton1998introduction} is that the reward signal is often hard to specify effectively \cite{skalse2024partial,miao2024inform}. Therefore, it often relies on learned reward signals that can be derived from demonstrations \cite{abbeel2004apprenticeship,ho2016generative}, preferences \cite{christiano2017deep}, intrinsic
objectives \cite{burda2018exploration,pathak2019self}, implicit feedback \cite{knox2009interactively}, safety constraints \cite{alshiekh2018safe,baert2025maximum}, or other
indirect supervision \cite{jin2020reward}. These sources are often easier to obtain than dense
scalar rewards: a human, verifier, or perception system may identify an unsafe
or unacceptable behavior without specifying how all other behaviors should be
ranked. The resulting learned reward model, however, is only useful if it provides effective guidance for the agent during training.

This is not guaranteed, since a reward model can fit its training data while still assigning high reward to unwanted behavior. Policy optimization can exploit these errors because it searches for behaviors that the model scores highly. For example, when training a robot to walk using a reward model trained on walking demonstrations, the reward model might also reward an unusual stationary pose. A policy could then learn to remain in that pose instead of walking. This connects learned reward models to reward misspecification, reward hacking, and reward-model overoptimization~\cite{amodei2016concrete,pan2022effects,gao2023scaling}.

We study how to update learned rewards using external corrective judgments without retraining the reward model. These judgments may identify unwanted behavior that the model rewards, or introduce requirements absent from its training data. Our method, \emph{flywheel patching}, incorporates both through local reward suppression using an Alignment Flywheel \cite{malomgre2026alignment,malomgre2026interactionless}. The aim is to retain useful learned knowledge while incorporating corrections and changing requirements~\cite{malomgre2026the}. The patching mechanism compiles verified findings into patches that locally reduce the deployed reward. The learned reward model remains fixed, and only the reward mapping transforming representations into rewards is edited. Corrective judgments can come from users, monitors, or domain-specific verifiers. The same interface can also incorporate requirements that were not expressed in the data used to train the reward model.

We formulate this mechanism for Interactionless Inverse Reinforcement Learning (IIRL) \cite{malomgre2026interactionless}. In IIRL, reward models are constructed from offline information without requiring online interaction with the environment. Our concrete instantiation is Mixture of Experts Guidance using Unlabeled and Incomplete Data for Exploration (MoE-GUIDE) \cite{malomgre2025mixture}. It learns a mixture-of-experts (MoE) reconstruction model from state-only expert observations and converts reconstruction consistency into an intrinsic reward. Each expert encodes a state as a compact latent vector, and we place local reward patches in this latent space.

Earlier MoE-GUIDE experiments reported reward overgeneralization, with the reward model assigning high rewards to non-expert states \cite{malomgre2025mixture}. They also used a short expert-state reset curriculum, initializing early policy rollouts near expert states. Here, the reward model is still learned from demonstrations. Downstream Soft Actor-Critic (SAC) \cite{haarnoja2018soft} training instead starts from the standard environment reset distribution. The policy receives the learned intrinsic reward alongside the environment reward. This provides a stricter test of whether learned and patched IIRL rewards can guide policy optimization without expert-state warm starts.

This paper makes four contributions:
\begin{itemize}
    \setlength{\itemsep}{0pt}
    \setlength{\parsep}{0pt}
    \item We introduce flywheel-guided patching for IIRL reward models. It incorporates external corrections and new requirements through local reward suppression.
    \item We extend the Alignment Flywheel to MuJoCo with candidate generation, expert-envelope verification, MoE-GUIDE integration, and regression-tested patch construction.
    \item We retrieve local patches in MoE-GUIDE's latent space to reduce the cost of online reward evaluation.
    \item We evaluate expert-envelope patching in MuJoCo without expert-state resets. In separate experiments, we compare compiled operator-specified norms with direct runtime enforcement.
\end{itemize}

The main benchmark uses a simple verifier that flags states outside the
coordinate-wise range observed in the demonstrations. This is a
dataset-relative support check, not a general out-of-distribution detector
or a solution to reward hacking. We also perform norm experiments using operator-specified
rules that are absent from or conflict with the demonstrations. Both supply
corrective judgments through the same patching interface, which can also
accept input from other monitors or human review.

\section{Related Work}
\label{sec:related_work}

Several research areas address parts of the problem studied here. Model editing
asks how a deployed model can be corrected without retraining it from scratch.
Reward-model and reward-hacking research studies what happens when an agent
exploits imperfections in the reward signal it is optimizing. Constraint and shielding
methods instead enforce requirements during learning or action selection. Our
method combines a different aspect of these ideas: it locally modifies the
reward signal seen by a reinforcement-learning agent, using external evidence
about where that reward should change.

The closest technical precedent for our method is lifelong model editing.
General Retrieval Adaptors for Continual Editing (GRACE) maintains a
frozen base model and attaches local corrections in
representation space: incoming examples are matched against stored keys, and
the corresponding edit changes the model output without updating the original
weights~\cite{hartvigsen2023aging}. Subsequent editing work has explored
related mechanisms for preserving unrelated behavior, including
null-space-constrained edits that project updates away from preserved knowledge
\cite{fang2025null,liu2026betaedit}, dynamic or side-memory approaches to
lifelong editing~\cite{wang2024wise}, and patcher-style methods that associate
localized errors with compact corrective modules~\cite{huang2023transformer}.
These methods motivate the abstraction used here: deployed models can be
maintained by accumulating local feature-space corrections rather than by
retraining from scratch.

Our setting differs in what we edit. Prior model-editing methods patch predictive behavior; we patch a reward surface optimized by a downstream RL policy. Policy optimization actively searches for high-reward behaviors and can exploit errors in this surface. This connects our setting to reward misspecification, reward hacking, and reward-model overoptimization \cite{amodei2016concrete,pan2022effects,gao2023scaling}. Reward editing can also introduce requirements that were absent from, or conflict with, the training data. Such updates need not correct a model error. Our patches therefore do not replace a prediction with a new label. They locally suppress the deployed reward to reduce incentives for behaviors that users or verifiers identify as unwanted or constraint-violating.

Another way to handle an unwanted behavior is to prevent it directly.
Constrained reinforcement learning and shielding do this by checking or
enforcing requirements while the policy learns or acts
\cite{achiam2017constrained,alshiekh2018safe}. Flywheel-guided patching takes a
different approach: it changes the reward signal as a soft correction mechanism, not a formal safety shield.
Demonstrations, safety events, operator-specified norms, human review, or
verifier outputs need only identify where the learned reward should change,
rather than define a complete replacement reward function.

A related alternative is to keep corrective checks outside the reward model
and apply them at runtime. Shielding methods, for example, monitor and
restrict the agent's actions~\cite{alshiekh2018safe}. This is often sensible for a cheap
deterministic rule. It becomes more costly when the judgment requires a
simulator, perception model, learned verifier, or human review.
Flywheel-guided patching instead uses such information during patch construction
and deploys only the resulting local reward modification. It therefore trades
the strength and immediacy of runtime enforcement for the ability to reuse a
corrective judgment without repeatedly running the original verifier.

\section{Interactionless Inverse Reinforcement Learning Artifacts}
\label{sec:iirl}

MoE-GUIDE separates the learned evaluator from the reward mapping. The evaluator is a mixture-of-experts model trained to reconstruct state-only expert observations. The mapping converts its reconstruction error into an intrinsic reward \cite{malomgre2025mixture}. Flywheel patching modifies this mapping without retraining the evaluator. Interactionless Inverse Reinforcement Learning (IIRL) extends this approach to a broader family of reward models constructed from offline information \cite{malomgre2026interactionless}. We use IIRL for the general class and MoE-GUIDE as the concrete model evaluated here.

We write the base reward as
\[
r_0(s) = g(E_\theta(s),s),
\]
where $E_\theta$ is the trained evaluator and $g$ is the explicit mapping that
turns the evaluator's outputs into a scalar intrinsic reward.  In MoE-GUIDE,
$E_\theta$ is a mixture-of-experts autoencoder and $g$ maps expert assignment,
bottleneck structure, and reconstruction consistency into reward.

A verified correction is compiled into a patched mapping
\[
g_{\mathcal{P}},
\]
where $\mathcal{P}$ is the set of accepted flywheel patches.  The patched reward
is
\[
r_{\mathcal{P}}(s)=g_{\mathcal{P}}(E_\theta(s),s).
\]

Reward-model overgeneralization is one motivation for patching. A learned evaluator may assign high intrinsic reward to states outside the intended expert-support region. Downstream policy optimization can exploit these errors by searching for high-reward behavior. Patching can also incorporate new requirements that are absent from or conflict with the demonstrations.

\section{Using the Alignment Flywheel for Reward Correction}
\label{sec:flywheel_mujoco}

The Alignment Flywheel provides the process around the patching mechanism. In
simple terms, it separates searching for possible reward failures from
verifying them, constructing a correction, and checking that the correction
does not damage behavior that should be preserved
\cite{malomgre2026alignment}. The corrective signal does not need to specify a
complete replacement reward. It only needs to identify a state or region where
the current reward should change.

To look for potential flaws in the reward model, we search through an adaptive sampling distribution
over the state space. Regions around available data receive higher initial
priority. When a flaw is discovered, probability mass is increased around that
region so that nearby states are examined more closely. Conversely, regions
that have already been explored are down-weighted, shifting search effort
toward less examined parts of the space. Candidate states produced by this
adaptive search are then passed to the Verifier. The Red Team therefore does
more than test a fixed set of states: it allocates search effort according to
what has already been observed and what earlier searches have uncovered.

We use this process with two kinds of corrective information. In the main MuJoCo benchmark, an expert-envelope verifier marks highly rewarded states outside the coordinate-wise range observed in the demonstrations. This simple check flags possible reward overgeneralization relative to the available data. It is not a general out-of-distribution detector and does not establish that these states are unsafe. In the norm experiments, the same interface receives operator-specified rules. These include wall clearance, path proximity, lane discipline, and distance from other agents in the controlled setting. Such requirements may be absent from or conflict with the demonstrations used to train the reward model.

For each verified finding, the Flywheel proposes a localized correction,
regression-tests it against protected expert behavior, and commits only
accepted patches. The reported experiments use cheap programmatic verifiers.
The same interface could in principle receive findings from slower sources such
as simulation, perception models, or human review, but those richer verifier
pipelines are not evaluated here. Section~\ref{sec:method} describes how
accepted findings are compiled into the deployed reward mapping.

\section{Flywheel-Guided Suppression Patching}
\label{sec:method}

\begin{figure}
    \centering
    \includegraphics[width=1\linewidth]{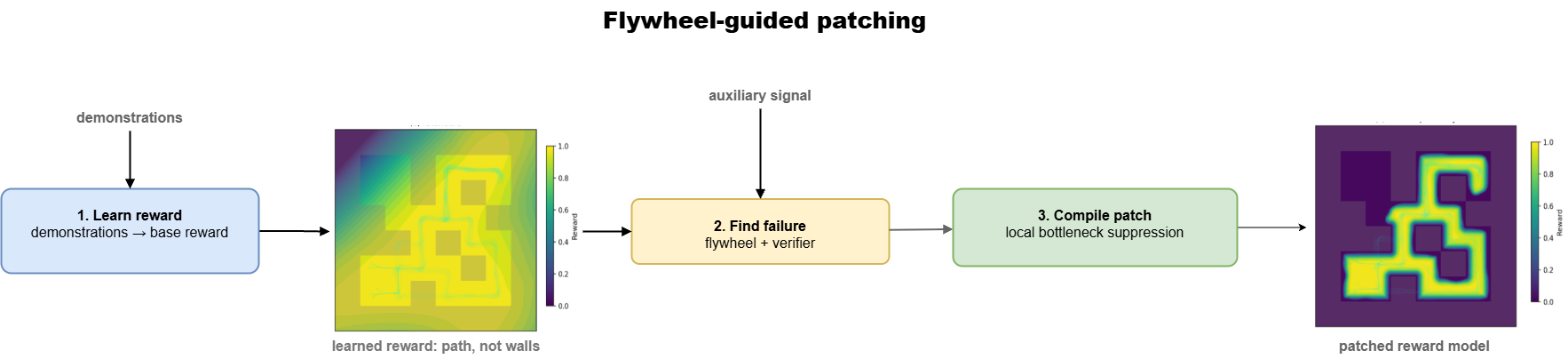}
    \caption{Flywheel-guided patching. Demonstrations train a base IIRL reward artifact.
    The Alignment Flywheel uses external judgments to identify regions where
    reward should be reduced, either to address overgeneralization or to
    incorporate new requirements. Verified corrections are compiled into local
    suppression patches in latent space. These patches modify the reward
    mapping without retraining the learned evaluator.}
    \label{fig:flywheel_guided_patching}
\end{figure}

The overview of the pipeline of the method can be seen in Figure \ref{fig:flywheel_guided_patching}. The method has a simple goal: keep the learned evaluator, but change the reward
it produces in regions that have been identified as problematic. First, an
external verifier identifies states that should not continue receiving high
reward. Second, the Alignment Flywheel converts each accepted finding into a
local suppression patch. During later reinforcement learning, the original
reward is reduced only when the current state is sufficiently close to a
patched region. The remainder of this section explains how failures are
identified, how patches are represented, and how they are evaluated efficiently
online.

The central intervention is to extend the explicit reward-mapping function $g$.
In the base artifact, a fixed learned evaluator $E_\theta$ produces
reward-relevant features and $g$ maps them to an intrinsic reward,
\[
r_0(s)=g(E_\theta(s),s).
\]
Given a flywheel-accepted patch set $\mathcal{P}$, we deploy instead
\[
r_{\mathcal{P}}(s)
=
g_{\mathcal{P}}(E_\theta(s),s)
=
\operatorname{clip}_{[0,r_{\max}]}
\left(
r_0(s)-\lambda S_{\mathcal{P}}(s)
\right).
\]
Thus, patching does not update the parameters of $E_\theta$. It updates the
reward artifact by replacing $g$ with $g_{\mathcal{P}}$, where
$S_{\mathcal{P}}$ is a flywheel-constructed suppression field. This is the sense
in which external corrective information updates the reward model: constraints,
support checks, perception signals, or other auxiliary diagnostics are not used
as rewards themselves, but are compiled into the mapping that produces reward.

\subsection{Reward-Free Corrective Signals}
\label{sec:reward_free_failure_signals}

Patching starts from an external judgment about where reward should be
reduced. This judgment may flag possible reward overgeneralization or
introduce a new requirement. We call this a corrective signal. It identifies
candidate patch targets without specifying a complete scalar reward function.

The Alignment Flywheel uses these signals to select regions for patching.
In the main MuJoCo benchmark, the signal comes from an expert-envelope
verifier derived from the demonstrations. Let $\mathcal{D}_E$ be the expert
state dataset and define the coordinate-wise bounds
\[
\ell_d=\min_{s\in\mathcal{D}_E}s_d,
\qquad
u_d=\max_{s\in\mathcal{D}_E}s_d .
\]
We define the out-of-support (OOS) flag for states outside this envelope as:
\[
\operatorname{OOS}(s)
=
\mathbf{1}
\left[
\exists d:\; s_d < \ell_d \;\lor\; s_d > u_d
\right].
\]
These are the base min/max bounds. Patch-construction variants can
expand them using the strictness-dependent margins defined in
Appendix~\ref{app:constraint_definitions}. In those variants, the expanded
bounds replace $\ell_d$ and $u_d$ in the check above.

If a flagged state also receives high intrinsic reward, the flywheel
marks it as a candidate for suppression. This is a simple, dataset-relative
signal of possible reward overgeneralization. It does not establish that
the state is unsafe or that its reward is necessarily incorrect.

In the norm experiments, corrective signals come from operator-specified
rules. These include wall clearance, path proximity, lane discipline, and
distance from other agents in the controlled setting. Such requirements
may be absent from or conflict with the demonstrations. A rule violation
can therefore motivate a patch even when the reward model accurately
reflects its training data.

The same interface can accept perception checks, simulation-based
diagnostics, or human review. These signals are used during patch
construction to decide where $g$ should change. The deployed reward
mapping then applies the resulting patches without rerunning the
original checks at every environment step.

\subsection{Local Suppression in Latent Space}
\label{sec:bottleneck_suppression}

Once a corrective judgment is accepted, we need to decide where reward
should be reduced. Editing only the observed state would be too narrow.
A global change could affect behavior we want to preserve. We therefore
place a local patch around the selected state in MoE-GUIDE's learned
latent space.

Each expert $e$ maps a state $s$ to a latent vector $z_e(s)$ and receives
a gating weight $\alpha_e(s)$. These vectors provide compact coordinates
for placing patches. Locality here refers to distance in the learned
representation.

A patch is a record
\[
p_j=(e_j,c_j,\sigma_j,\eta_j,\tau_j,m_j),
\]
where $e_j$ is the associated expert, $c_j$ is the latent-space center,
$\sigma_j$ is the kernel bandwidth, $\eta_j$ is the suppression strength,
$\tau_j$ records the patch type, and $m_j$ stores provenance metadata from
the flywheel. The center is the latent vector of the state selected for
correction, typically under the expert with the highest gating weight.

Let
\[
\mathcal{P}_e = \{p_j \in \mathcal{P}: e_j=e\}
\]
denote the accepted patches assigned to expert $e$. The exact suppression
field defined by the patch set is
\[
S_{\mathcal{P}}(s)
=
\min\left(
1,\;
\sum_{e\in\mathcal{E}}
\alpha_e(s)
\sum_{p_j\in\mathcal{P}_e}
\eta_j
\exp\left(
-\frac{\|z_e(s)-c_j\|^2}{2\sigma_j^2}
\right)
\right),
\]
where $\mathcal{E}$ is the set of MoE experts. Each patch contributes
less as the latent vector moves away from its center. The outer minimum
caps the combined suppression at one.

The patched reward is
\[
r_{\mathcal{P}}(s)
=
g_{\mathcal{P}}(E_\theta(s),s)
=
\operatorname{clip}_{[0,r_{\max}]}
\left(
r_0(s)-\lambda S_{\mathcal{P}}(s)
\right),
\]
where $r_0(s)=g(E_\theta(s),s)$ and $\lambda$ controls the scale of
suppression. The patches modify the reward mapping while leaving the
learned evaluator $E_\theta$ unchanged.

Before commitment, each candidate patch is checked against protected
demonstration states. Patches that reduce reward on these states are
rejected or narrowed. The corrective signal identifies where reward
should be reduced. The regression check limits changes to behavior
designated for preservation.

\subsection{Scalable Online Evaluation}
\label{sec:scalable_patch_eval}

The exact suppression field in Section~\ref{sec:bottleneck_suppression}
sums over all accepted patches. Checking every patch at each environment
step becomes more expensive as patches accumulate. We therefore retrieve
nearby candidates in latent space and compute suppression only for this
subset.

For each expert $e$, the patch centers in $\mathcal{P}_e$ are stored in a
locality-sensitive hash index over that expert's bottleneck codes. At reward
evaluation time, the current code $z_e(s)$ is hashed using the same projections,
and only patches in the matching or neighboring buckets are considered. Let
\[
\mathcal{C}_e(s)=\operatorname{Retrieve}(z_e(s),\mathcal{P}_e)
\]
be the retrieved candidate set for expert $e$. The deployed approximation to
the suppression field is
\[
\widehat{S}_{\mathcal{P}}(s)
=
\min\left(
1,\;
\sum_{e\in\mathcal{E}}
\alpha_e(s)
\sum_{p_j\in\mathcal{C}_e(s)}
\eta_j
\exp\left(
-\frac{\|z_e(s)-c_j\|^2}{2\sigma_j^2}
\right)
\right).
\]
The reward used by the downstream policy optimizer is therefore
\[
r_{\mathcal{P}}(s)
=
\operatorname{clip}_{[0,r_{\max}]}
\left(
r_0(s)-\lambda \widehat{S}_{\mathcal{P}}(s)
\right).
\]

This changes the online cost from scaling with the full patch set,
$O(\sum_e|\mathcal{P}_e|)$, to scaling with the number of retrieved local
candidates, $O(|\mathcal{E}|C)$, where $C$ is the average number of candidates
retrieved per expert. In the reported experiments, we evaluate all experts,
because the MoE artifacts have few experts. For larger reward artifacts, the
same retrieval scheme can be combined with top-$K$ expert pruning by evaluating
only the experts with the largest gating weights. This reduces the cost to
$O(K_E C)$, but this additional pruning is a scaling extension rather than part
of the reported experimental protocol.

The expert-envelope verifier used in the main MuJoCo benchmark is cheap enough to evaluate at every step. For cheap deterministic rules, direct runtime checking may be simpler and can achieve stronger compliance. Compilation becomes more attractive when judgments are expensive or unavailable during control. These may come from perception models, simulation-based diagnostics, formal checkers, or human review. The flywheel uses them during patch construction to identify where reward should be reduced. They can address reward errors or introduce new requirements. At deployment, applying these corrections requires only local patch retrieval and suppression. Our experiments test reward editing and its performance–compliance trade-offs using cheap rules. Evaluating the computational benefit with expensive verifiers remains future work.

\subsection{Downstream Policy Optimization}
\label{sec:downstream_policy_optimization}

After patch construction, the agent uses the patched reward without any
change to its RL algorithm. At each environment step, we compute the
original intrinsic reward and apply suppression from the retrieved patches.
The resulting intrinsic reward is
\[
r_{\mathcal{P}}(s)=g_{\mathcal{P}}(E_\theta(s),s).
\]

We use Soft Actor-Critic (SAC) \cite{haarnoja2018soft} for policy training.
In the main MuJoCo benchmark, the patched intrinsic reward is added to the
environment reward. The reward model is trained from demonstrations, but
policy rollouts start from the standard environment reset distribution.
Unlike the earlier MoE-GUIDE evaluation, this benchmark uses no
expert-state reset curriculum. The separate sparse Ant norm experiments
retain that curriculum.

We evaluate patches through their effect on learned behavior. The main
benchmark tests corrections based on the expert envelope. The norm
experiments test requirements absent from or conflicting with the
demonstrations. Runtime-enforcement baselines check the rule directly
at each step instead of applying precomputed patches.

\section{Results}
\label{sec:results}

Table~\ref{tab:main-results} shows undiscounted evaluation returns.
Each method is trained with five seeds. Policies are evaluated every
10K environment steps using 10 episodes per seed. We average episode
returns within each seed, then report the mean $\pm$ standard deviation
across the five seed averages. The reported checkpoints are 500K steps
for Swimmer and 1M steps for all other environments. Appendix results
use the same aggregation unless stated otherwise.

The baselines cover several alternatives to patching. Soft Actor-Critic
(SAC) uses only the environment reward (ER). ER+RND adds a Random Network
Distillation (RND) \cite{burda2018exploration} exploration bonus, while ER+Disagreement  adds a bonus
based on disagreement between predictive models~\cite{pathak2019self}. These test whether
standard exploration bonuses explain the gains. f-IRL \cite{ni2021f} provides an
inverse reinforcement learning baseline and also receives environment
reward during training. Boundary Enforcement checks the expert-envelope
rule directly at each step. MoE-GUIDE supplies the unpatched reward model.
Flywheel Patching uses the same learned evaluator and incorporates
external corrective judgments through the patched mapping
$g_{\mathcal{P}}$.

\subsection{Compiling constraints and operator-specified norms}
\label{sec:norms}

\begin{table}[t]
\centering
\caption{Compiled patches versus online enforcement of the same rule, in
both settings. Upper: controlled quadruped, task throughput at matched
compliance ($8$ agents, $8$ seeds), read off each swept front by
interpolation. Lower: MuJoCo sparse Ant at the last available checkpoint
region, five seeds, expert-state reset curriculum retained. Higher
return is better; lower violation is better.}
\label{tab:norm_front}
% AUTHOR CHECK: The sparse-Ant any-violation values are also reported in
% tab:ant_full, where they are smaller than the hip-violation component.
% Resolve that numerical/aggregation inconsistency before final submission.
\small
\begin{tabular}{lcc}
\toprule
\multicolumn{3}{l}{\emph{Controlled quadruped} -- throughput at matched collisions} \\
Collisions per agent & $0.90$ & $0.60$ \\
\midrule
Compiled patch        & $\mathbf{3.83}$ & $\mathbf{3.62}$ \\
Runtime, formal rule  & $3.03$ & $2.87$ \\
Runtime, graded rule  & $3.63$ & $3.29$ \\
\midrule
\multicolumn{3}{l}{\emph{MuJoCo sparse Ant}} \\
Method & Return & Any violation \\
\midrule
Unpatched MoE-GUIDE          & $2.89$ & $88.4\%$ \\
Compiled kernel, combined    & $\mathbf{5.22}$ & $80.2\%$ \\
Runtime OOB, hard            & $0.20$ & $\mathbf{1.9\%}$ \\
Runtime OOB, smooth          & $1.72$ & $\mathbf{1.6\%}$ \\
\bottomrule
\end{tabular}
\end{table}

Before reporting the benchmark returns, we ask two simpler questions:
can the same patching interface compile an operator-specified rule that was not
present in the demonstrations, and when is compilation preferable to checking
that rule online at every step? We use two settings. A controlled planar
quadruped isolates the mechanism because norm compliance has closed-form ground
truth at every step. MuJoCo sparse Ant then tests the same comparison where the
constraint conflicts with both the demonstration data and the task reward.

\paragraph{Controlled setting: norms the demonstrations never express.}
The main benchmark compiles the expert-envelope
verifier, which is derived from the demonstrations. Here we compile four
norms an operator writes down: keep $0.13$ from any wall, stay near the
demonstrated route, travel on the right-hand side of a corridor relative
to heading, and keep a set distance from other agents. The agent-distance rule uses a channel appended to the observation.
This channel is available to the verifier but is not an input to the
learned reward model.
% AUTHOR CHECK: Does this distance channel affect patch-target selection,
% runtime patch evaluation, or both? Specify the actual implementation here.
The reward artifact is a
mixture-of-experts autoencoder over a $23$-dimensional observation of
which only two dimensions are position and sixteen are joint state, so a
positional norm must be written into a surface that is mostly a function
of gait. Demonstrations are norm-neutral by construction -- mean wall
clearance $0.40$ with $1.9\%$ of states below threshold, and $50.6\%$ of
corridor states on the right against a chance rate of $50\%$ -- so the
compiled patches cannot sit on the demonstrated route and the lane
asymmetry cannot be inherited from the data.

The norms compile and they bind. Over free space, the share of
still-rewarded states too close to a wall falls from $18.9\%$ to
$6.0\%$ and off the demonstrated path from $4.1\%$ to $0.0\%$, and
corridors acquire a $5.71\times$ right-over-left reward asymmetry where
they had $1.00\times$. Downstream, lane occupancy rises from $50\%$ to
$98\%$ and collisions fall from $1.90$ to $0.27$ per agent. Compared
conditions compile \emph{identical} patches and differ only in the
enforcement weight, with weight $0$ meaning ``compiled but not
enforced'', so every effect is attributable to enforcement rather than
to the conditions holding different reward models.

\paragraph{Task throughput at matched compliance.}
% AUTHOR CHECK: Define task throughput here, including the quantity counted,
% its units or normalization, and the evaluation interval.
We sweep the enforcement weight to compare task throughput at different collision rates. Table~\ref{tab:norm_front} compares compiled patches with formal and graded runtime versions of the same rule. All three reduce collisions. At matched collision rates, compiled patches achieve higher task throughput in this controlled setting. The graded runtime rule narrows this gap. Patches concentrate suppression around targets accepted by the flywheel. Runtime rules apply penalties wherever the rule is violated. This difference may help patches preserve task progress, although the comparison does not isolate the contribution of locality. The formal runtime rule achieves the lowest observed collision rate: $0.209$ collisions per agent, compared with $0.271$ for compiled patches. At these operating points, task throughput is $1.87$ for the formal rule and $2.27$ for the patches.

Enforcement is also not uniformly a tax. For the agent-distance norm,
moderate enforcement raises final success from $0.82$ to $0.90$,
plausibly because independent learners in a shared corridor face
less non-stationarity when they collide less often. The three positional
norms cost sample efficiency monotonically in weight; lane discipline
reaches $96\%$ compliance at a weight where no seed reaches $50\%$
success.

\paragraph{Sparse Ant: constraints that conflict with the demonstrations.}
In sparse Ant, roughly $78\%$ of the training states violate at least one
constraint. Each constraint is violated in $30$--$35\%$ of demonstration
states. Forward progress can also earn reward while constraints are
violated. The new rules therefore conflict with both the demonstrations
and the task incentives.

The lower block of Table~\ref{tab:norm_front} shows the results. Compiled
patches reduce violations from $88.4\%$ to $80.2\%$ while increasing
return from $2.89$ to $5.22$. Hard runtime out-of-bounds (OOB) enforcement
reduces violations to $1.9\%$, but return falls to $0.20$. Smooth runtime
enforcement achieves a similar violation rate of $1.6\%$ with a higher
return of $1.72$.

The hard-rule policy barely moves. Its low violation rate therefore does
not show that it can satisfy the constraints while performing the task.
These runs retain the expert-state reset curriculum, so poor performance
persists even with rollouts initialized near demonstrated states. The
results are exploratory and cover only part of training. Training extent
and additional comparisons are reported in
Appendix~\ref{app:norms:ant}.

Graded runtime enforcement also preserves more task performance in the
controlled setting. At $0.90$ collisions per agent, it achieves throughput
of $3.63$, compared with $3.03$ for the formal rule. Both settings therefore
show better task performance with graded enforcement at similar compliance.

\paragraph{When to use compiled patches.}
The two settings illustrate different trade-offs. In the controlled
setting, compiled patches preserve more task throughput at matched
collision rates. In sparse Ant, the constraints conflict with the
demonstrations. Patching improves return but reduces violations only
modestly. Runtime enforcement reduces violations much further, but
also lowers return. For cheap deterministic rules, direct runtime checking may be simpler,
especially when reducing violations is the priority. Compilation becomes
more attractive when judgments are expensive or unavailable at every
step, as with simulation, perception, or human review. These judgments
can be used during patch construction and reused through local reward
edits. Our experiments demonstrate performance--compliance trade-offs
using cheap rules. They do not establish a computational advantage
for expensive verifiers; evaluating that benefit remains future work.
Full setup for both settings, per-norm fronts, learning curves,
trajectory visualizations and reproducibility detail are in
Appendix~\ref{app:norms}.

\subsection{Downstream benchmark performance}
\label{sec:benchmark_results}
\begin{figure}[H]
    \centering
    \includegraphics[width=1\linewidth]{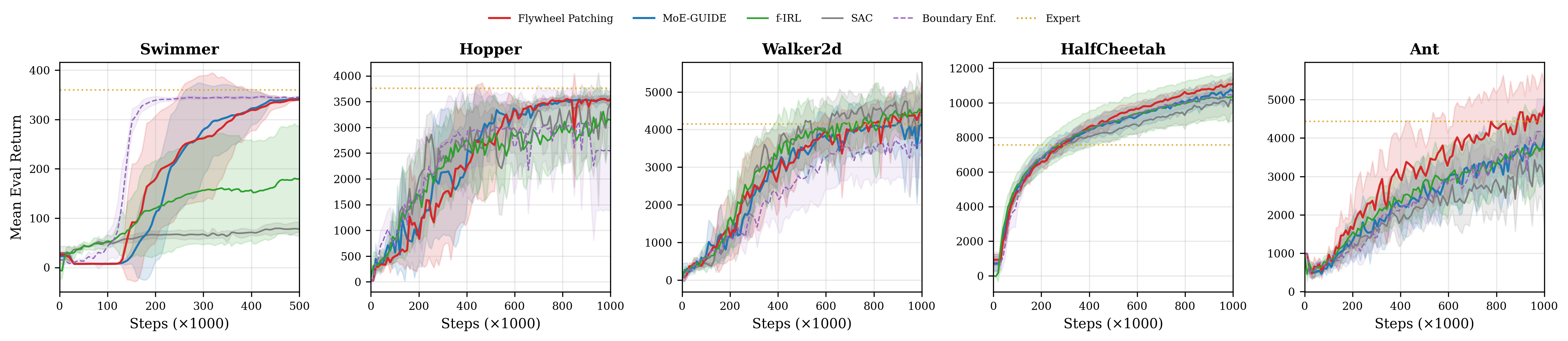}
    \caption{Learning curves accompanying the main results in Table \ref{tab:main-results}}
    \label{fig:main_curves}
\end{figure}

Table~\ref{tab:main-results} and Figure~\ref{fig:main_curves} show the downstream benchmark results.
Flywheel Patching achieves the highest mean return in three of five
environments and a higher mean than unpatched MoE-GUIDE in four.
The largest increase over MoE-GUIDE is in Ant, at 842 return points,
followed by HalfCheetah at 445 and Walker2d at 389. Mean returns change
little in Hopper and Swimmer. The largest gains therefore occur in
the three environments with imperfect demonstrations. The baselines show that several approaches can improve downstream
learning. Exploration bonuses produce substantial mean gains over SAC
in Ant. f-IRL achieves the highest mean in Walker2d and a similar mean
to Flywheel Patching in HalfCheetah. Boundary Enforcement achieves the
highest mean in Swimmer, while Flywheel Patching has higher means in
the other four environments. These results show that local reward
patches can improve downstream performance while retaining the same
learned evaluator.

\begin{table}[]
\centering
\caption{Mean undiscounted evaluation return $\pm$ standard deviation
across five seeds, with 10 evaluation episodes averaged per seed.
Swimmer is evaluated at 500K steps; all other environments at 1M steps.
Bold marks the highest sample mean among the compared learning methods,
excluding the expert reference. Methods differ in their use of demonstrations, exploration
bonuses, and corrective checks.
$^\ddagger$Imperfect expert demonstrations.}
\label{tab:main-results}
\small
\begin{tabular}{@{}lccccc@{}}
\toprule
Method & Swimmer & Hopper & Walker2d$^\ddagger$ & HalfCheetah$^\ddagger$ & Ant$^\ddagger$ \\
\midrule
Expert           & 360  & 3{,}761 & 4{,}150 & 7{,}582 & 4{,}443 \\
\midrule[0.01pt]
SAC              & $78 \pm 15$    & $3459 \pm 127$  & $4353 \pm 916$  & $10125 \pm 657$ & $2807 \pm 1149$ \\
ER+RND           & $98 \pm 29$    & $3181 \pm 330$  & $4450 \pm 778$  & $10492 \pm 529$ & $4018 \pm 771$  \\
ER+Disagreement  & $71 \pm 11$    & $3240 \pm 261$  & $4445 \pm 466$  & $10359 \pm 282$ & $3812 \pm 866$  \\
\midrule[0.01pt]
f-IRL            & $329 \pm 17$   & $3464 \pm 91$   & $\mathbf{4580 \pm 474}$ & $10996 \pm 496$ & $3454 \pm 663$ \\
MoE-GUIDE        & $342 \pm 2$    & $3540 \pm 86$   & $4146 \pm 877$  & $10646 \pm 657$ & $3974 \pm 1099$ \\
\midrule[0.01pt]
Boundary Enf.    & $\mathbf{352 \pm 2}$ & $3349 \pm 109$ & $3052 \pm 1085$ & $9711 \pm 626$ & $2687 \pm 1241$ \\
Flywheel Patching & $341 \pm 3$ & $\mathbf{3544 \pm 76}$ & $4535 \pm 722$ & $\mathbf{11091 \pm 307}$ & $\mathbf{4816 \pm 594}$ \\
\bottomrule
\end{tabular}
\end{table}

The baselines show that several approaches can improve downstream
learning. Exploration bonuses produce substantial mean gains over SAC
in Ant. f-IRL achieves the highest mean in Walker2d and a similar mean
to Flywheel Patching in HalfCheetah. Boundary Enforcement achieves the
highest mean in Swimmer, while Flywheel Patching has higher means in
the other four environments. These results show that local reward
patches can improve downstream performance while retaining the same
learned evaluator.

Auxiliary experiments are summarized in Table~\ref{tab:aux-summary}
and detailed in the appendix. They examine patch placement, retrieval
cost, and operator-specified constraints. At coverage levels of $75\%$ and above, latent-space ball covers use
fewer centers than raw-state covers in the reported comparison.
This finding supports their use in the tested setting, but does not
establish that the representation is suitable for every corrective
rule. Hashed retrieval restricts suppression evaluation to retrieved
candidates. The constraint experiments extend the evaluation beyond
expert-envelope corrections and examine how task reward affects
compliance.

\begin{table}[t]
\centering
\caption{Auxiliary findings in the tested settings. Detailed results
are reported in
Appendices~\ref{app:scalable_eval}--\ref{app:constraint_results}.}
\label{tab:aux-summary}
\small
\begin{tabular}{lp{0.63\linewidth}}
\toprule
Experiment & Finding \\
\midrule
Latent vs. raw-state patches &
Latent-space ball covers use fewer centers than raw-state covers
at coverage levels of $75\%$ and above. \\

Hashed retrieval &
Suppression is evaluated only for retrieved candidates. Kernel
evaluation cost depends on the number of candidates returned. \\

Constraint-targeted Ant &
Patches reduce violations under intrinsic-reward-only evaluation.
Violation reductions are weaker when environment reward is included. \\

Runtime OOB enforcement &
Direct runtime enforcement produces larger violation reductions in
Ant, but can also reduce task return. \\
\bottomrule
\end{tabular}
\end{table}

Together, the experiments show how external support checks and
operator-specified rules can be incorporated into a learned reward
without retraining its evaluator.

\section{Discussion and Conclusion}
\label{sec:discussion_conclusion}

Flywheel patching updates a learned reward through local edits to its
explicit reward mapping. Accepted corrective judgments become suppression
patches in $g_{\mathcal{P}}$, while the learned evaluator $E_\theta$
remains fixed. The same mechanism can address reward overgeneralization
and incorporate new requirements. Corrective information need only
identify where reward should be reduced; it does not need to define
a complete replacement reward.

In the main MuJoCo benchmark, policy training starts from standard
environment resets without expert-state warm starts. Patching achieves
higher mean returns than unpatched MoE-GUIDE in four of five environments.
The largest increases occur in Ant, HalfCheetah, and Walker2d.
The norm experiments show that the same interface can incorporate
requirements absent from or conflicting with the demonstrations.
In the controlled setting, compiled patches preserve more task throughput
at matched collision rates. In sparse Ant, patching improves return with
a modest reduction in violations, while runtime enforcement produces
stronger violation reductions at lower returns.

\paragraph{Limitations.}
The main benchmark uses five seeds, with substantial variation on
several tasks. More runs would help assess smaller differences between
methods. The evaluation also relies on MoE-GUIDE and its learned latent
space. We have not established how well the method transfers to other
reward models or representations. The expert-envelope verifier checks
only coordinate-wise demonstration bounds, so it provides a limited
signal of possible overgeneralization. Patching is a soft reward edit and does not guarantee constraint
satisfaction. The runtime reward penalties tested here do not provide
such guarantees either. Direct runtime checking may be simpler for
cheap rules, especially when reducing violations is the priority.
Compilation becomes more attractive when judgments are expensive or
unavailable at every step. Our experiments use cheap verifiers and
do not establish this computational benefit.

\paragraph{Future work.}
Future work should evaluate richer corrective signals, including human
review, learned support models, perception checks, and simulator
diagnostics. Further directions include longer patch sequences,
automatic patch parameter selection, and applications beyond MoE-GUIDE.
These include other intrinsic-reward methods and reward machines
\cite{bellemare2016unifying,burda2018exploration,aubret2023information,icarte2018using,icarte2022reward,baert2025reward,shehab2026active}.

Code for MoE-GUIDE, Alignment Flywheel, and flywheel patching is available
at \url{https://github.com/decide-ugent/Alignment-Flywheel}.\section*{Acknowledgment}
This research was supported by funding from the Flemish Government under the ‘‘Onderzoeksprogramma Artificiele Intelligentie (AI) Vlaanderen’’ program.

\appendix

\section{Compiled Norms: Setup and Full Results}
\label{app:norms}

This appendix gives the full setup and results behind
Section~\ref{sec:norms}. The controlled planar quadruped isolates the mechanism
under norms with known ground truth, while MuJoCo sparse Ant tests the same
comparison when the requested constraint conflicts with the demonstration data
and task reward.

%% =====================================================================
\subsection{Controlled quadruped: environment and artifact}
\label{app:norms:setup}

The body is a planar quadruped walking a serpentine corridor layout of
$1\times1$ cells. Locomotion is not scripted at the torso: a central
pattern generator drives eight joints, feet in stance exert ground
friction, and the torso moves only under the resulting forces. Steering
is differential, so a commanded heading is realised by shortening the
stride on the inside of the turn and the body must yaw before it can
translate. The action set is eight headings at full and $0.4$ stride
plus a halt; without a throttle the body has a single speed, and a
following-distance norm is then unsatisfiable in a queue.

The observation is $23$-dimensional: position $(2)$, heading $(2)$,
planar velocity and yaw rate $(3)$, and eight joints with their rates
$(16)$. Demonstrations come from a scripted controller that descends a
geodesic to the goal while paying to stay clear of walls, with $12\%$
random actions so runs spread across the corridor width. A
mixture-of-experts autoencoder is fitted to those runs and its
reconstruction error mapped to reward exactly as elsewhere in this
paper. Queries into the artifact are planar: position, heading and twist
are written from the query and the sixteen joint dimensions copied from
the nearest demonstrated state, so the model is only ever asked about
gait poses the body genuinely adopted.

\begin{table}[h]
\centering
\caption{What the artifact pays for, before and after compiling the four
norms, over a grid of free space at the demonstrated speed. Percentages
are over the states the model still rewards, so the columns measure what
the reward function endorses rather than how much space it covers.}
\label{tab:norm_surface}
\small
\begin{tabular}{lrrrr}
\toprule
& Rewarded & Too close & Off the & On the \\
& space & to wall & path & left side \\
\midrule
OOD only  & $100.0\%$ & $18.9\%$ & $4.1\%$ & $45.5\%$ \\
All norms & $73.5\%$  & $\mathbf{6.0\%}$ & $\mathbf{0.0\%}$ & $\mathbf{35.3\%}$ \\
\bottomrule
\end{tabular}
\end{table}

Table~\ref{tab:norm_surface} shows the compiled surface. Corridors
additionally acquire a $5.71\times$ right-over-left reward asymmetry
where they previously had $1.00\times$, from $387$ positional, $477$
lane and $108$ crowding patches.

%% =====================================================================
\subsection{Sweep design}
\label{app:norms:sweep}

Each cell is one $(\text{norm}, \text{mechanism}, \text{weight},
\text{separation}, \text{seed})$ combination trained for $300$ episodes
with $8$ agents in two-way corridor traffic and evaluated greedily. The
grid covers four norms, three enforcement mechanisms, eight enforcement
weights, four separation thresholds for the agent-distance norm, and
eight seeds: $1{,}248$ runs.
% AUTHOR CHECK: Reconcile this run count with the 832 cells reported in
% app:norms:limits; state explicitly if the latter is an audited subset.
Weight $0$ is the control -- patches
compiled, enforcement off.

Two parameters were set by observed failure rather than chosen a priori,
and both are properties of the body rather than of the method. First,
the separation threshold must exceed the distance the body covers in one
decision: at the point mass's $0.35$ the crowding norm does essentially
nothing at four agents ($0.71$ against a $0.72$ baseline), because a
quadruped at the demonstrated speed covers about $0.35$ of a cell per
decision and must yaw before it can move sideways, so the norm first
registers one decision before contact. At $0.50$ the identical patches
remove two thirds of the collisions. Second, the episode horizon must
leave room for a compliant route: the point-mass protocol allows a
scripted controller roughly $4.5\times$ the time it needs, and matching
that ratio for this body requires $700$ steps rather than $320$.

%% =====================================================================
\subsection{Per-norm fronts and learning curves}
\label{app:norms:fronts}

\begin{figure}[h]
\centering
\includegraphics[width=\linewidth]{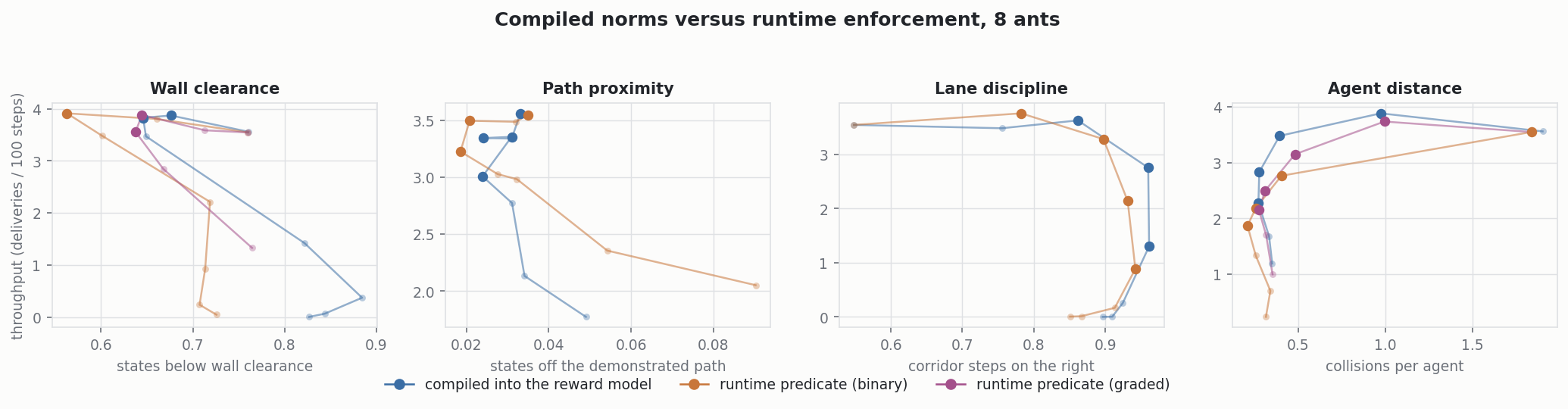}
\caption{Compliance against task throughput for each norm, $8$ agents,
$8$ seeds per point. Each point is one enforcement weight applied to
identical compiled patches; larger markers are Pareto-optimal.}
\label{fig:norm_front}
\end{figure}

Beyond the agent-distance result in the main text, lane discipline
reaches $0.96$ compliance under compilation against $0.55$ unenforced,
with compiled dominating runtime on throughput at matched compliance;
path proximity has little room to move, since the OOD surface already
holds agents near the demonstrated route ($3.4\%$ off-path unenforced);
and wall clearance is weak for every mechanism at this density, with
runtime marginally better than compilation on that norm alone. With
eight agents in a corridor one unit wide the bodies are pushed against
walls, and the metric counts corners where clearance is inherently low.

\begin{figure}[h]
\centering
\includegraphics[width=\linewidth]{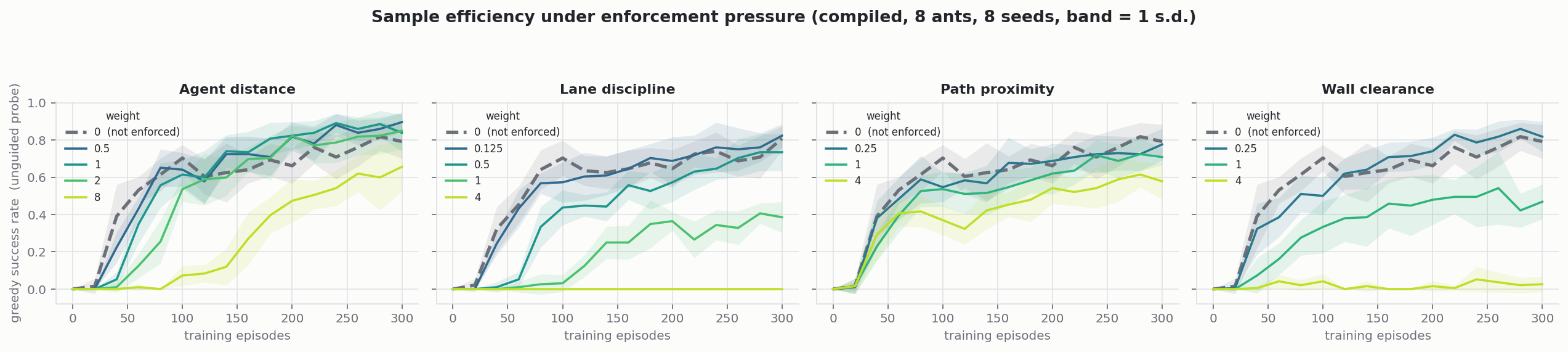}
\caption{Unguided greedy-probe success against training episodes, $8$
seeds, band $=1$ s.d. Dashed grey is the control: identical patches,
enforcement off. Moderate enforcement of the agent-distance norm raises
the ceiling; the three positional norms cost sample efficiency
monotonically in weight.}
\label{fig:norm_curves}
\end{figure}

%% =====================================================================
\subsection{Trajectories}
\label{app:norms:traj}

\begin{figure}[h]
\centering
\includegraphics[width=\linewidth]{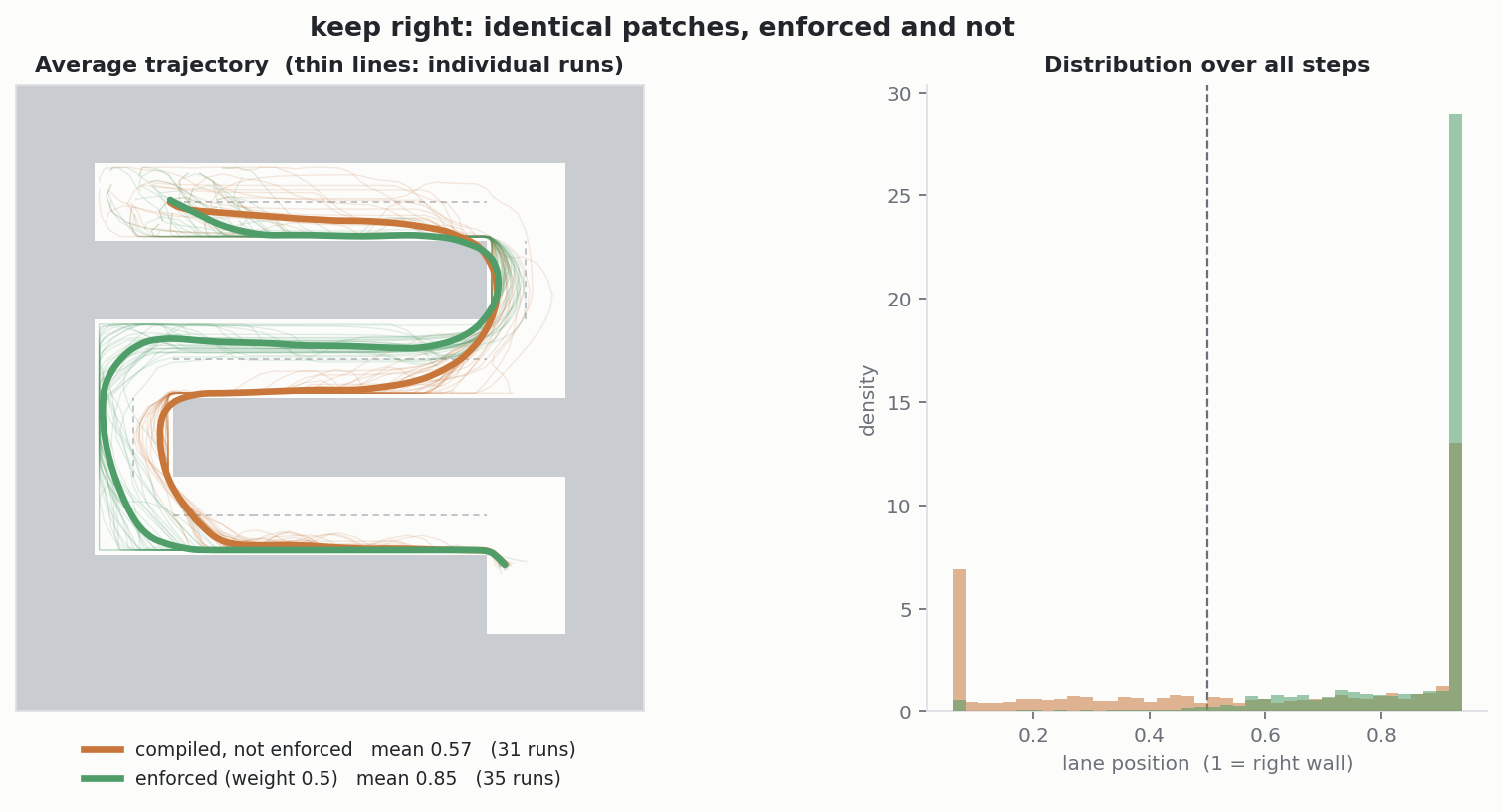}
\caption{Average forward trajectory per condition over the faded
individual runs, with the distribution of lane position beside it. Only
forward-running agents are drawn: ``right'' is relative to heading, so
drawing both directions of two-way traffic places compliant traffic
against both walls and cancels the effect visually. Both conditions
compile identical patches and differ only in enforcement weight.}
\label{fig:norm_lane_traj}
\end{figure}

\begin{figure}[h]
\centering
\includegraphics[width=\linewidth]{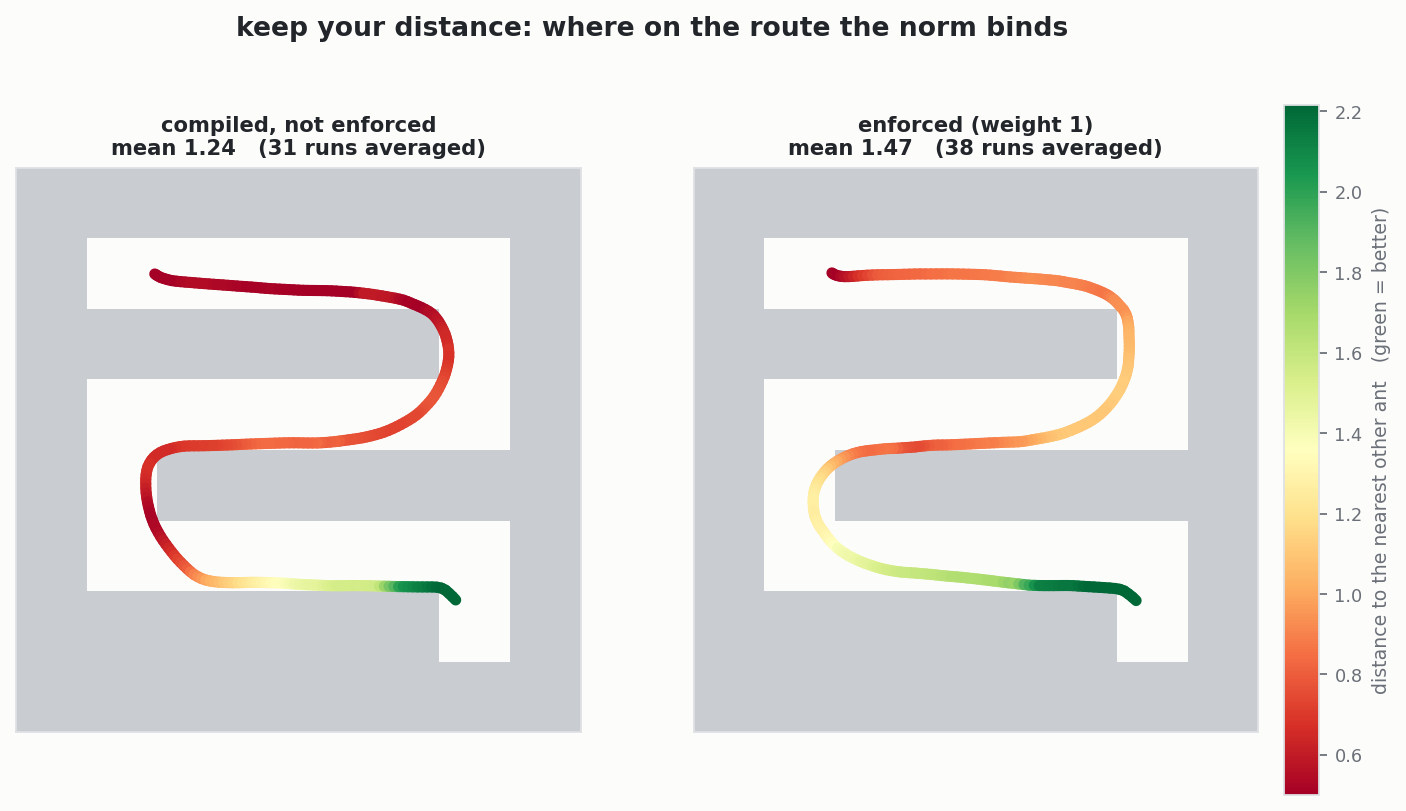}
\caption{The average route coloured by mean distance to the nearest
other agent. Red marks where agents habitually bunch up, which is where
collisions come from; enforcement shifts the whole route away from the
crowded end.}
\label{fig:norm_distance_heat}
\end{figure}

%% =====================================================================
\subsection{Sparse Ant: protocol and full comparison}
\label{app:norms:ant}

The sparse Ant environment awards $+10$ whenever the torso crosses the
next forward checkpoint, while constraint signals measure speed, torso
tilt and hip velocity. Forward progress remains profitable in a
constraint-violating state, so performance and constraint satisfaction
must be considered jointly.

Two protocol facts matter for interpretation. These runs retain the
expert-state reset curriculum for learning stability, unlike the main
results of Table~\ref{tab:main-results}, which deliberately remove it;
and most runs reached approximately $480$--$490$K of the intended $1$M
environment steps, so they should be read as exploratory partial-training
evidence rather than completed benchmark results.

\begin{table}[h]
\centering
\caption{Sparse Ant at the last available checkpoint region. Descriptive
means over five-seed trajectories; no standard deviations or confidence
intervals.}
\label{tab:ant_full}
% AUTHOR CHECK (NUMERICAL): Any violation must be at least each component
% rate when computed on the same transitions with the same aggregation.
% The hard row shows 1.9 percent any versus 12.9 percent hip; the smooth
% row shows 1.6 percent any versus 13.5 percent hip. Verify the source
% values and denominators before updating this table, tab:norm_front,
% and the related approximately-2-percent claims in the main text.
\small
\begin{tabular}{lrrr}
\toprule
Method & Return & Any violation & Hip velocity \\
\midrule
Unpatched MoE-GUIDE            & $2.89$ & $88.4\%$ & $56.3\%$ \\
Combined kernel, medium        & $\mathbf{5.22}$ & $80.2\%$ & $45.4\%$ \\
Hip-velocity kernel, very tight & $3.30$ & $70.4\%$ & $41.4\%$ \\
OOB hard $+$ combined safety   & $0.20$ & $\mathbf{1.9\%}$ & $12.9\%$ \\
OOB smooth $+$ combined safety & $1.72$ & $\mathbf{1.6\%}$ & $13.5\%$ \\
\bottomrule
\end{tabular}
\end{table}

A performance-matched reading is also available: over the $30$ common
checkpoints where the very-tight hip-velocity kernel matches or exceeds
the unpatched baseline, mean return is $3.69$ against $2.59$, with
any-constraint violations $75.9\%$ against $88.3\%$, hip-velocity
$45.3\%$ against $57.9\%$ and tilt $50.9\%$ against $57.1\%$. Because
the subset is selected by performance this is descriptive rather than an
unbiased test, and we report the unselected last-checkpoint values in
Table~\ref{tab:ant_full} alongside it.

%% =====================================================================
\subsection{Limitations and reproducibility}
\label{app:norms:limits}

The controlled body is a planar approximation: a knee coordinate stands
in for foot height rather than a vertical axis being simulated, so the
gait is a top-down trot, and the $23$-dimensional surface is queried
through the planar adapter described above. The claim supported is that
norm compilation survives a reward model whose input is mostly
\emph{not} position; it is not a claim about compiling norms that are
themselves statements about gait.

The sweep was run twice with identical code and identical seeds,
distributed differently across a heterogeneous cluster. $389$ of $832$
cells returned bitwise identical results and $443$ did not. A controlled
test isolates the cause: the same cell run twice on one Rome node gives
$1.604167$ collisions both times, and twice on one Turin node gives
$2.000000$ both times -- deterministic on a machine, different between
machines. Independent learners in a shared environment are chaotic, so
floating-point differences between CPU microarchitectures are enough to
separate otherwise identical runs. A seed alone therefore does not pin a
result here, the hardware being part of the configuration, and the
eight-seed means across the two sweeps differ by about $4\%$ while the
single-cell cross-architecture spread is much larger. All reported
figures and tables are drawn from a single sweep.

\section{Reward artifact and Training Protocol}
\label{app:artifact_protocol}

This appendix documents the reward artifact, downstream training protocol, baselines, patch-construction details, scaling measurements, and full auxiliary analyses supporting the main results. The main paper keeps the mechanism compact; the appendix provides the complete reproducibility details. Unless stated otherwise, training-result tables report mean evaluation return $\pm$ standard deviation across seeds using the same checkpoint evaluation protocol as the main text.

\subsection{MoE Reward Model Architecture}
\label{app:moe_architecture}

Each expert in the Mixture of Gated Autoencoders is a single-hidden-layer autoencoder: a linear encoder ($d_\text{obs} \to d_\text{bottleneck}$) with ReLU activation, followed by a linear decoder ($d_\text{bottleneck} \to d_\text{obs}$). The gating network is a linear layer with softmax output over experts. Table~\ref{tab:moe-architecture} lists the per-environment model dimensions. All models are pre-trained on $N_\text{traj}$ expert demonstration trajectories.

\begin{table}[h]
\centering
\caption{MoE reward model architecture per environment.}
\label{tab:moe-architecture}
\small
\begin{tabular}{@{}lccccc@{}}
\toprule
Environment & $d_\text{obs}$ & $d_\text{bottleneck}$ & Experts & $N_\text{traj}$ & Expert type \\
\midrule
Swimmer-v2      & 8   & 3  & 1 & 5 & Perfect \\
Hopper-v2       & 11  & 4  & 2 & 5 & Perfect \\
Walker2d-v2     & 17  & 7  & 4 & 5 & Imperfect \\
HalfCheetah-v2  & 17  & 7  & 4 & 5 & Imperfect \\
Ant-v2          & 111 & 10 & 3 & 5 & Imperfect \\
\bottomrule
\end{tabular}
\end{table}

For expert $e$, the encoder and decoder are
\begin{align}
  \mathbf{z}_e &= \operatorname{ReLU}(W_e^{\text{enc}}\,\mathbf{x}
                   + \mathbf{b}_e^{\text{enc}}),
                   & W_e^{\text{enc}} \in \mathbb{R}^{B \times D},\;
                     \mathbf{b}_e^{\text{enc}} \in \mathbb{R}^B, \\
  \hat{\mathbf{x}}_e &= W_e^{\text{dec}}\,\mathbf{z}_e
                        + \mathbf{b}_e^{\text{dec}},
                   & W_e^{\text{dec}} \in \mathbb{R}^{D \times B},\;
                     \mathbf{b}_e^{\text{dec}} \in \mathbb{R}^D.
\end{align}
The gating network is a single linear layer followed by softmax,
\begin{equation}
  \alpha_e(\mathbf{x}) = \frac{\exp(w_e^\top \mathbf{x} + b_e^g)}
                         {\sum_{j=1}^E \exp(w_j^\top \mathbf{x} + b_j^g)},
  \qquad w_e \in \mathbb{R}^D,\; b_e^g \in \mathbb{R}.
\end{equation}
The gating-weighted reconstruction is $\hat{\mathbf{x}} = \sum_{e=1}^{E} \alpha_e(\mathbf{x})\,\hat{\mathbf{x}}_e$.

\subsection{Intrinsic Reward Mapping}
\label{app:reward_mapping}

The intrinsic reward is computed by an exponential mapping from per-observation reconstruction loss to $[0,1]$:
\begin{equation}
  g(L) = \mathrm{clip}\!\left(\exp\!\left(-s \cdot \frac{L - L_\text{min}}{L_\text{max} - L_\text{min}}\right),\;0,\;1\right),
\end{equation}
where $L$ is the per-observation MSE reconstruction loss. In downstream training, the total reward is $r = r_\text{ext} + \kappa_t \cdot g(L)$ with $\kappa_t = \kappa_{t-1}\cdot\kappa_\text{decay}$ applied at each SAC update step. Table~\ref{tab:reward-mapping} lists the per-environment mapping parameters used in the main results.

\begin{table}[h]
\centering
\caption{Intrinsic reward mapping parameters per environment. These are the best hyperparameters selected via the $L_\text{max}$ sweep.}
\label{tab:reward-mapping}
\small
\begin{tabular}{@{}lcccccc@{}}
\toprule
Environment & $L_\text{max}$ & $L_\text{min}$ & Steepness $s$ & $\kappa$ & $\kappa_\text{decay}$ & Steps \\
\midrule
Swimmer-v2      & 0.2   & 0.11  & 25  & 1 & 1.0       & 500K \\
Hopper-v2       & 0.15  & 0.13  & 25  & 2 & 1.0       & 1M \\
Walker2d-v2     & 0.6   & 0.07  & 100 & 2 & 0.9999995 & 1M \\
HalfCheetah-v2  & 1.0   & 0.5   & 100 & 1 & 0.999995  & 1M \\
Ant-v2          & 0.1   & 0.003 & 100 & 1 & 0.9999995 & 1M \\
\bottomrule
\end{tabular}
\end{table}

\subsection{SAC and Replay Buffer Settings}
\label{app:sac_settings}

All environments use the same SAC configuration, listed in Table~\ref{tab:sac-settings}. Swimmer uses a 100K-transition replay buffer; all other environments use 1M. Transitions store the combined reward $r_\text{total}=r_\text{ext}+\kappa_t\cdot g(L)$ at insertion time.

\begin{table}[h]
\centering
\caption{SAC hyperparameters, shared across all environments.}
\label{tab:sac-settings}
\small
\begin{tabular}{@{}lr@{}}
\toprule
Parameter & Value \\
\midrule
Actor/Critic hidden layers      & $[256, 256]$ \\
Actor/Critic activation         & ReLU \\
Learning rate (all networks)    & $3 \times 10^{-4}$ \\
Discount factor $\gamma$        & 0.99 \\
Polyak averaging $\tau$         & 0.995 \\
Automatic entropy tuning        & Yes \\
Target entropy                  & $-d_\text{act}$ \\
Batch size                      & 256 \\
Random exploration steps        & 10{,}000 \\
Evaluation interval             & 10{,}000 steps \\
Evaluation episodes per checkpoint & 10 \\
Number of seeds                 & 5 \\
\bottomrule
\end{tabular}
\end{table}

\section{Baseline Implementations}
\label{app:baselines}

\subsection{Flywheel Patching Variant Used in the Main Table}
\label{app:fk-best-variant}

The Flywheel Patching row in Table~\ref{tab:main-results} uses the best-performing kernel variant per environment, selected from the tightness and patch-identification ablations (Tables~\ref{tab:tightness} and~\ref{tab:patch-method}). Table~\ref{tab:fk-variants} lists the specific variant used in each environment. HalfCheetah and Ant use the original kernel, because the unmodified flywheel patching achieved the highest return in those environments.

\begin{table}[h]
\centering
\caption{Flywheel Patching variant used per environment in Table~\ref{tab:main-results}.}
\label{tab:fk-variants}
\small
\begin{tabular}{@{}lll@{}}
\toprule
Environment & Tightness & Patch method \\
\midrule
Swimmer-v2      & Loose      & Regression \\
Hopper-v2       & Tight      & Regression \\
Walker2d-v2     & Tight      & Regression \\
HalfCheetah-v2  & (original) & (original) \\
Ant-v2          & (original) & (original) \\
\bottomrule
\end{tabular}
\end{table}

\subsection{Exploration Baselines: ER+RND and ER+Disagreement}
\label{app:exploration-baselines}

Both exploration baselines use the SAC configuration in Table~\ref{tab:sac-settings} with an additive intrinsic reward $r = r_\text{ext}+\lambda r_\text{int}$, where $r_\text{int}$ is recomputed at sample time using the current module state.

\paragraph{RND.} Random Network Distillation uses a fixed random target network and a trainable predictor, both $d_\text{obs}\to64\to64\to64$ MLPs with ReLU activations and LayerNorm on the output. The intrinsic reward is $r_\text{int}=\frac{1}{d}\sum_i (f_\theta(s)-f_{\bar\theta}(s))_i^2$. The predictor is updated with Adam (lr $=10^{-3}$) at every timestep.

\paragraph{Disagreement.} Ensemble Disagreement uses an ensemble of 5 forward dynamics models, each $(d_\text{obs}+d_\text{act})\to256\to256\to d_\text{obs}$ with ReLU activations. The intrinsic reward is the variance of next-state predictions across ensemble members, averaged over observation dimensions. The ensemble is trained with Adam (lr $=10^{-3}$) at every timestep.

\paragraph{Scale sweep.} Both methods are evaluated with $\lambda\in\{0.001,0.01,0.1\}$ for each environment. The scale yielding the highest mean return at evaluation time, 500K for Swimmer and 1M for all other environments, is selected for Table~\ref{tab:main-results}. All runs use 2M total timesteps, a 1M replay buffer, 10K random exploration steps, and 5 seeds.

\subsection{f-IRL Baseline}
\label{app:firl-baseline}

We use the original f-IRL implementation with the reverse KL divergence objective. f-IRL alternates between training a discriminator on expert versus agent state distributions, optimizing a reward function with the $f$-divergence loss, and running SAC with the learned reward.

\paragraph{Reward network.} The reward network is an MLP with ReLU activations and output clamped to $[-10,10]$. Hidden sizes are $[64,64]$ for all environments except Ant, which uses $[128,128]$. It is trained with Adam (lr $=10^{-4}$, weight decay $=10^{-3}$), one gradient step per IRL iteration.

\paragraph{Discriminator.} The discriminator is ResNet-style with tanh activations: 3 residual blocks ($d_\text{hid}=128$) for all environments except Ant, which uses 6 blocks. It is trained with Adam (lr $=3\times10^{-4}$, weight decay $=10^{-4}$) for 1{,}200 steps per IRL iteration with batch size 800.

\paragraph{SAC integration and scale sweep.} f-IRL uses the same SAC architecture and hyperparameters as MoE-GUIDE. At each SAC update, the reward is $r=r_\text{ext}+\lambda r_{f\text{-}\mathrm{IRL}}(s)$. Each environment is swept over $\lambda\in\{0.01,0.1,1.0\}$ with 5 seeds. The best scale per environment is: Swimmer $=1.0$, Hopper $=0.1$, Walker2d $=0.1$, HalfCheetah $=0.01$, Ant $=0.01$. Swimmer uses 500K steps; all other environments use 1M steps, matching the MoE-GUIDE training budget.

\subsection{Boundary and OOB Enforcement Baselines}
\label{app:boundary_baselines}

Out-of-bounds enforcement checks active observation dimensions against the demonstrated $[\min,\max]$ range. Hard mode clips intrinsic reward to 0 when any dimension is out of bounds. Smooth mode applies a Gaussian decay penalty. Optional safety functions add constraint-specific checks. Full OOB and constraint-targeted results are reported in Section~\ref{app:constraint_results}.

\section{Flywheel Patch Construction Details}
\label{app:patch_construction}
\label{app:flywheel_implementation}

\subsection{Expert-Envelope and Constraint Definitions}
\label{app:constraint_definitions}

The implementation expands the base min/max envelope from
Section~\ref{sec:reward_free_failure_signals} using strictness-dependent
margins. At strictness $s\in[0,1]$,
\begin{align}
  \text{lo}_d &= \mu_d^{\min} - (\mu_d^{\text{mean}} - \mu_d^{\min})(1-s), \\
  \text{hi}_d &= \mu_d^{\max} + (\mu_d^{\max} - \mu_d^{\text{mean}})(1-s),
\end{align}
where $\mu_d^{\min}$, $\mu_d^{\max}$, and $\mu_d^{\text{mean}}$ are per-dimension demonstration statistics. At $s=1$, the bounds equal the observed minima and maxima. The margin is asymmetric: each side shrinks proportionally to its own half-range. The named levels are \texttt{very\_loose} ($s=0$), \texttt{loose} ($s=0.2$), \texttt{mild} ($s=0.4$), \texttt{medium} ($s=0.6$), \texttt{tight} ($s=0.8$), and \texttt{very\_tight} ($s=1$).

% AUTHOR CHECK: This paragraph uses tilt 46 degrees and hip velocity 8,
% whereas app:constraint-setup uses 45 degrees and 5. Identify which
% experiment uses each definition, or correct the parameters from the code.
For Ant, three explicit physical constraints are used: velocity $\sqrt{v_x^2+v_y^2}\leq 6.0$, torso tilt $2\arccos(|q_w|)\leq 46^\circ$, and hip velocity $\max_{i\in\{19,21,23,25\}}|v_i|\leq8.0$. The combined constraint is violated if any of these checks fails. Each constraint returns a boolean violation mask and an out-of-bounds ratio $\rho=\mathrm{clip}((v-\theta)/\theta,0,\infty)$.

\subsection{Strictness and Patch Parameters}
\label{app:strictness_params}

Each strictness level controls patch placement through a reward floor,
a regression tolerance, and a bandwidth range. For demonstration-preserving
patches, $\Delta_{\max}=0$ requires that accepted patches do not reduce reward
on the demonstration basin. Constraint-targeted patches that intentionally
suppress violating demonstrations are treated separately in
Section~\ref{app:constraint_results}.

\begin{table}[h]
\centering
\caption{Strictness parameters for the Flywheel governance loop.}
\label{tab:strictness}
\small
\begin{tabular}{lrrrr}
\toprule
Level & $\tau_{\text{floor}}$ & $\Delta_{\max}$ & $\sigma_{\min}$ & $\sigma_{\max}$ \\
\midrule
\texttt{very\_loose} & 0.70 & 0.0 & 0.50 & 4.0 \\
\texttt{loose}       & 0.50 & 0.0 & 0.40 & 3.5 \\
\texttt{medium}      & 0.30 & 0.0 & 0.30 & 3.0 \\
\texttt{tight}       & 0.15 & 0.0 & 0.20 & 2.5 \\
\texttt{very\_tight} & 0.05 & 0.0 & 0.15 & 2.0 \\
\texttt{adaptive}    & 0.05 & 0.0 & 0.15 & 4.0 \\
\bottomrule
\end{tabular}
\end{table}

For fixed strictness, the bandwidth for a flaw with severity $\rho$ is
\begin{equation}
  \sigma = \mathrm{clip}\left(\sigma_{\min}+(\sigma_{\max}-\sigma_{\min})\min(\rho,1),\;\sigma_{\min},\;\sigma_{\max}\right).
\end{equation}
For adaptive strictness, the mapping is inverted:
\begin{equation}
  \sigma = \mathrm{clip}\left(\sigma_{\max}-(\sigma_{\max}-\sigma_{\min})\min(\rho,1),\;\sigma_{\min},\;\sigma_{\max}\right).
\end{equation}
Strength is $\mathrm{clip}(0.5+0.5\min(\rho,1),0.5,1.0)$ for fixed strictness and $\mathrm{clip}(0.5+0.45\min(\rho,1),0.5,0.95)$ for adaptive strictness.

\subsection{Flywheel Governance Loop}
\label{app:flywheel_loop}

The Alignment Flywheel wraps the reward artifact as the Oracle and decomposes correction into Red Team, Verifier, Triage, and Refinement roles. The Red Team searches the current reward surface for out-of-support states that remain highly rewarded. The Verifier confirms whether each candidate is a true violation. Triage prioritizes verified candidates. Refinement computes bandwidths and strengths, regression-tests candidate patches, and commits accepted corrections. All decisions are recorded in an append-only knowledge base.

\begin{algorithm}[h]
\caption{Flywheel Kernel Construction with Demonstration Preservation}
\label{alg:flywheel}
\begin{algorithmic}[1]
\REQUIRE Demo observations $\mathcal{D}$, adaptive candidate sampler $Q$, MoE model $M$, constraint function $C$, strictness parameters $(\tau_{\text{floor}},\Delta_{\max},\sigma_{\min},\sigma_{\max})$, patience $p$
\ENSURE Kernel $\mathcal{K}$ with patch set $\bigcup_e \mathcal{P}_e$
\STATE Initialise oracle with $M$ and $\mathcal{D}$ as basin
\STATE $\text{stale}\gets0$, $\text{best}\gets0$
\REPEAT
  \STATE \textbf{Red Team:} Sample candidate states $\mathcal{X}$ using the adaptive state-space sampler $Q$
  \STATE Find $\mathcal{F}=\{x\in\mathcal{X}\mid r_{\text{moe}}(x)>\tau_{\text{floor}} \wedge C(x)=\textsc{violated}\}$
  \STATE \textbf{Verify:} For each $x\in\mathcal{F}$, confirm violation and compute $\rho(x)$
  \FOR{each verified flaw $(x,\rho)$}
    \STATE Encode $x$ into bottleneck space for the gating-selected expert $e$
    \STATE Propose $(\mathbf{c}=\mathbf{z}_e,\sigma(\rho),s(\rho))$
    \STATE Shrink bandwidth while demo-basin reward loss exceeds $\Delta_{\max}$
    \IF{the shrunk patch satisfies the regression constraint}
      \STATE Commit patch $(\mathbf{c},\sigma,s)$ to $\mathcal{P}_e$
    \ELSE
      \STATE Reject patch
    \ENDIF
  \ENDFOR
  \STATE Update early stopping using flaw-count improvement
\UNTIL{$\text{stale}\geq p$ or iteration $\geq100$}
\STATE Export $\mathcal{K}=(M,\bigcup_e\mathcal{P}_e,\text{metadata})$
\end{algorithmic}
\end{algorithm}

\subsection{Regression Testing}
\label{app:patching:regression}

For the demonstration-preserving protocol in Algorithm~\ref{alg:flywheel},
the regression test evaluates each candidate patch against the full
demonstration basin $\mathcal{D}$. For demo point $x_d$, the reward change is
\begin{equation}
  \Delta r(x_d)=\operatorname{supp}_{\text{new}}(x_d)-\operatorname{supp}_{\text{old}}(x_d).
\end{equation}
The maximum reward loss $\max_{x_d}\Delta r(x_d)$ must not exceed $\Delta_{\max}$. In this protocol $\Delta_{\max}=0$, so no demo point may lose reward. If the constraint is violated, the bandwidth is reduced until the test passes or $\sigma<\sigma_{\min}$, in which case the patch is rejected.

% AUTHOR CHECK: Specify the protected states, if any, and the regression
% procedure used for norm and constraint-targeted variants. Table~\ref{tab:patch-suppression}
% shows intentional reward reductions on violating demonstration states.

\section{Scalable Patch Evaluation}
\label{app:scalable_eval}

\subsection{Suppression Function}
\label{app:suppression_function}

Given patches $\mathcal{P}_e=\{(\mathbf{c}_k,\sigma_k,s_k)\}_{k=1}^{K_e}$ for expert $e$, the total suppression for observation $\mathbf{x}$ is
\begin{equation}
  \operatorname{supp}(\mathbf{x}) = \min\!\left(1,\sum_{e=1}^{E}
    \alpha_e(\mathbf{x}) \sum_{k=1}^{K_e}
    s_k \exp\!\left(-\frac{\|\mathbf{z}_e(\mathbf{x}) - \mathbf{c}_k\|^2}{2\sigma_k^2}\right)\right).
  \label{eq:supp_full}
\end{equation}
Any Gaussian term below $10^{-3}$ is clamped to zero. This Gaussian floor gives finite support in practice and enables locality-sensitive hashing.

\subsection{Vectorized Brute-Force Evaluation}
\label{app:bruteforce_suppression}

The vectorized implementation pre-transposes patch centers into $(E,B,K_{\max})$ arrays, precomputes $\|\mathbf{c}_k\|^2$ and $1/(2\sigma_k^2)$, encodes all experts for a batch of $N$ observations, and uses $\|a-b\|^2=\|a\|^2+\|b\|^2-2a^\top b$ to compute all Gaussian values. The aggregate suppression is $\min(1,\sum_e \alpha_e\sum_k G_{e,k})$. The cost is $O(N\cdot E\cdot K_{\max})$ per batch.

\subsection{LSH Build and Query}
\label{app:lsh_details}

For every expert, patch centers are stored in an approximate nearest-neighbor index over bottleneck space. Each hash table samples normalized Gaussian hyperplanes, projects patch centers, thresholds each bit by the median projection for that expert, and groups patch indices by integer hash code. The default settings are reported in Table~\ref{tab:lsh_defaults}.

At query time, the current bottleneck code is hashed with the same projections and thresholds. Candidates from matching buckets are deduplicated across tables. The implementation then gathers the relevant $(o,e,k)$ triples, evaluates the Gaussian suppression only for those triples, and scatters the weighted values back to observations using the corresponding gating weights.

\begin{table}[h]
\centering
\caption{LSH default hyperparameters.}
\label{tab:lsh_defaults}
% AUTHOR CHECK: A bit cap of B+2 gives 12 bits when B=10, but the scaling
% tables report 14--24 bits. Specify any benchmark override or distinguish
% requested from effective bits, based on the implementation.
\small
\begin{tabular}{ll}
\toprule
Parameter & Value \\
\midrule
Number of tables $T$ & 4 \\
Bits per table $b$ & auto-scaled, range 14--24 \\
Bit cap & $B+2$ \\
Gaussian floor & $10^{-3}$ \\
Gaussian cutoff radius & $3.72\,\sigma$ \\
RNG seed & 42 \\
\bottomrule
\end{tabular}
\end{table}

\paragraph{Optional top-$K$ expert pruning.} The reported experiments use all experts, because the number of experts is small. For larger artifacts, the same LSH query can be combined with top-$K$ expert pruning by evaluating only the experts with the largest gating weights, masking and renormalizing inactive experts, and flattening only the active observation--expert pairs. This option is a scaling extension, not part of the reported experimental protocol.

\subsection{Scaling Measurements}
\label{app:scaling_measurements}

The ablation study isolates the suppression step only; encoding and gating are excluded to measure pure scaling behavior. The default setup uses bottleneck dimension $B=10$, batch size $N=1000$, query vectors $\operatorname{ReLU}(\mathcal{N}(0,3))$, patch centers sampled uniformly from $[0,5]^B$, bandwidth $\sigma=0.3$, strength $s=0.8$, and median timing over 5 runs with 1 warmup.

\begin{table}[h]
\centering
\caption{Suppression time vs. patch count ($E=1$, $B=10$, $N=1000$).}
\label{tab:patch_scaling}
\small
\begin{tabular}{rrrrr}
\toprule
$K$ & Bits & Brute (ms) & LSH (ms) & Speedup \\
\midrule
100       & 14 &      1.1 &    0.3 &     3.7$\times$ \\
1{,}000   & 14 &     12.4 &    1.2 &    10.1$\times$ \\
10{,}000  & 18 &    175.2 &    2.5 &    71.4$\times$ \\
100{,}000 & 21 &  1{,}420 &    6.2 &   228$\times$   \\
1{,}000{,}000 & 24 & 19{,}227 & 15.2 & 1{,}263$\times$ \\
\bottomrule
\end{tabular}
\end{table}

\begin{table}[h]
\centering
\caption{Suppression time vs. expert count ($K_e=10{,}000$, $B=10$, $N=1000$).}
\label{tab:expert_scaling}
\small
\begin{tabular}{rrrrrr}
\toprule
$E$ & $K_{\text{total}}$ & Brute (ms) & LSH (ms) & Brute/$E$ & LSH/$E$ \\
\midrule
 1 &  10{,}000  &    158 &    4.8 & 158 & 4.8 \\
 4 &  40{,}000  &    858 &   15.6 & 215 & 3.9 \\
16 & 160{,}000  & 4{,}615 &  108 & 288 & 6.7 \\
32 & 320{,}000  & 8{,}384 &  215 & 262 & 6.7 \\
64 & 640{,}000  & 17{,}318 & 482 & 271 & 7.5 \\
\bottomrule
\end{tabular}
\end{table}

\begin{table}[h]
\centering
\caption{Optional top-$K$ scaling study: suppression cost vs. expert count ($K_{\text{top}}=2$, $K_e=10{,}000$, $B=10$, $N=1000$). This is a large-scale extension and is not used in the reported MuJoCo experiments.}
\label{tab:topk_scaling}
\small
\begin{tabular}{rrrrrrr}
\toprule
$E$ & Dense BF & Top-2 BF & Dense LSH & Top-2 LSH
    & $\frac{\text{BF}_{K}}{\text{BF}}$
    & $\frac{\text{LSH}_{K}}{\text{LSH}}$ \\
\midrule
 2 &    502 &  518 &   13.0 &  13.5 & 103\% & 104\% \\
 4 &  1{,}257 &  971 &   28.2 &  28.8 &  77\% & 102\% \\
 8 &  2{,}048 & 1{,}467 &  55.6 &  24.2 &  72\% &  44\% \\
16 &  4{,}319 &  822 &  120 &  36.3 &  19\% &  30\% \\
32 &  8{,}479 &  334 &  216 &  33.6 & 3.9\% &  16\% \\
64 & 18{,}090 &  287 &  432 &  36.9 & 1.6\% & 8.5\% \\
\bottomrule
\end{tabular}
\end{table}

\section{Patch Robustness and Variant Ablations}
\label{app:patch_ablations}

\subsection{FP Patch Budget Sweep}
\label{app:fk-cap-sweep}

\begin{figure}[H]
\centering
\includegraphics[width=\linewidth]{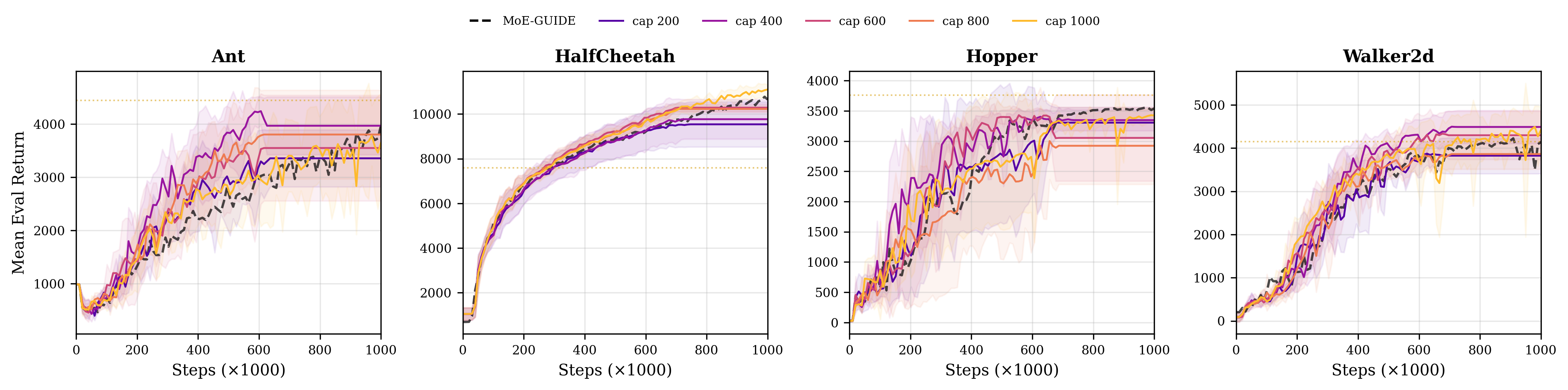}
\caption{Learning curves for the FP patch budget sweep in Table~\ref{tab:fk-cap-sweep}. Curves show mean evaluation return across seeds, with shaded bands for one standard deviation.}
\label{fig:fk-cap-learning-curves}
\end{figure}

To assess whether the flywheel kernel's patch budget affects task reward, we sweep the maximum number of patches from 200 to 1000 across four environments and compare against the unpatched MoE-GUIDE baseline. All variants share the same hyperparameters and differ only in the \texttt{--flywheel\_variant tight\_regression\_cap$N$} flag, which limits how many suppression patches the flywheel kernel may accumulate. Table~\ref{tab:fk-cap-sweep} and Figure~\ref{fig:fk-cap-learning-curves} shows the mean evaluation return at matched training steps using the available seed data. No comparison reaches statistical significance; point estimates trend positive in Ant and HalfCheetah and are mixed in Hopper and Walker2d.

\begin{table}[t]
\centering
\caption{FP patch cap sweep: mean eval return $\pm$ std. No comparison reaches significance (Welch's $t$-test, all $p>0.05$). $^*$Fewer than 5 seeds due to log parser failures.}
\label{tab:fk-cap-sweep}
\small
\begin{tabular}{@{}lcccc@{}}
\toprule
Method & \shortstack{Ant-v2\\(400K)} & \shortstack{HalfCheetah-v2\\(600K)} & \shortstack{Hopper-v2\\(600K)} & \shortstack{Walker2d-v2\\(600K)} \\
\midrule
MoE-GUIDE   & $2692 \pm 792$          & $9236 \pm 604$          & $3369 \pm 116$          & $3887 \pm 600$          \\
FP cap200   & $2639 \pm 799$          & $9303 \pm 879$          & $3371 \pm 221^*$        & $3535 \pm 449^*$        \\
FP cap400   & $3433 \pm 783$          & $9241 \pm 1003$         & $3354 \pm 122$          & $4367 \pm 316^*$        \\
FP cap600   & $2947 \pm 859$          & $9870 \pm 375$          & $3388 \pm 117$          & $4270 \pm 663^*$        \\
FP cap800   & $2995 \pm 618$          & $9637 \pm 164$          & $2417 \pm 1255$         & $3832 \pm 370^*$        \\
FP cap1000  & $2867 \pm 614$          & $9744 \pm 253$          & $2815 \pm 989$          & $3587 \pm 967$          \\
\bottomrule
\end{tabular}
\end{table}

\subsection{Constraint Tightness Ablation}
\label{app:tightness-ablation}

\begin{figure}[H]
\centering
\includegraphics[width=\linewidth]{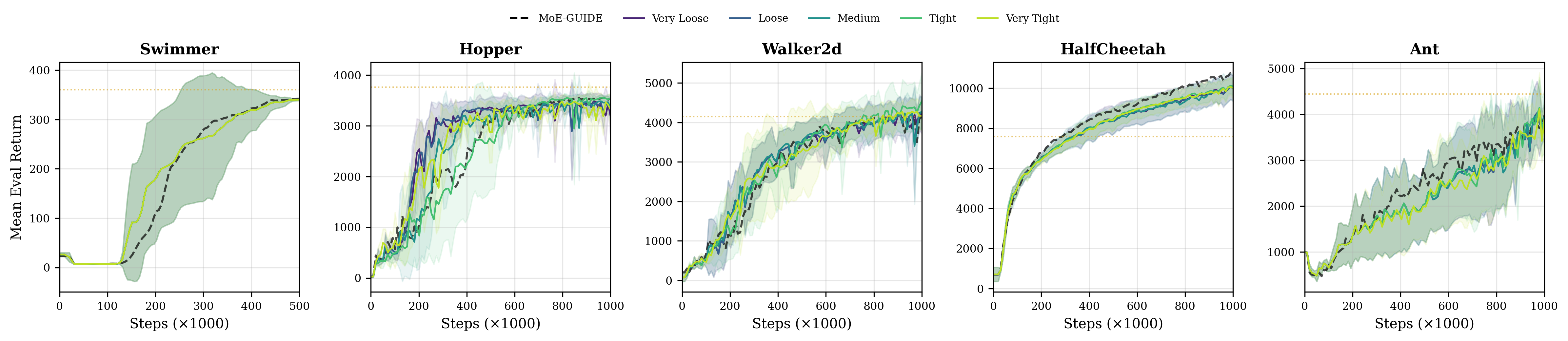}
\caption{FP constraint tightness ablation: final mean evaluation return ± std across
5 seeds. Swimmer at 500K steps; all others at 1M steps.}
\label{fig:tight-ablation}
\end{figure}

Table~\ref{tab:tightness} and Figure~\ref{fig:tight-ablation} show the effect of varying FP constraint tightness from very loose to very tight using the regression patch-identification method. MoE-GUIDE is shown as the baseline. Performance is largely robust to the tightness setting: variations are within one standard deviation in most environments, suggesting that the method is not highly sensitive to this hyperparameter.

\begin{table}[h]
\centering
\caption{FP constraint tightness ablation: final mean evaluation return $\pm$ std across 5 seeds. Swimmer at 500K steps; all others at 1M steps.}
\label{tab:tightness}
\small
\begin{tabular}{@{}lccccc@{}}
\toprule
Method & Swimmer & Hopper & Walker2d & HalfCheetah & Ant \\
\midrule
MoE-GUIDE    & $342 \pm 2$   & $3540 \pm 86$   & $4146 \pm 877$  & $10646 \pm 657$ & $3974 \pm 1099$ \\
\midrule
Very Loose   & $339 \pm 3$   & $3191 \pm 425$  & $4174 \pm 457$  & $10004 \pm 587$ & $3858 \pm 366$  \\
Loose        & $341 \pm 3$   & $3335 \pm 198$  & $4259 \pm 405$  & $10010 \pm 661$ & $3858 \pm 366$  \\
Medium       & $339 \pm 3$   & $3344 \pm 264$  & $4280 \pm 374$  & $9980 \pm 627$  & $3941 \pm 527$  \\
Tight        & $339 \pm 3$   & $3544 \pm 76$   & $4535 \pm 722$  & $10097 \pm 457$ & $3827 \pm 681$  \\
Very Tight   & $340 \pm 2$   & $3322 \pm 318$  & $4172 \pm 354$  & $10016 \pm 555$ & $3875 \pm 403$  \\
\bottomrule
\end{tabular}
\end{table}

\subsection{Patch-Identification Method: Regression vs. DBSCAN}
\label{app:patch-method}

Table~\ref{tab:patch-method} compares two patch-identification strategies for the kernel at tight strictness: linear regression of reconstruction error on observation dimensions and DBSCAN clustering of high-error observations. Both methods yield similar final returns across environments, with regression performing slightly better in Walker2d and DBSCAN showing a notable drop in Swimmer.

\begin{table}[h]
\centering
\caption{Regression vs. DBSCAN patch identification: final mean evaluation return $\pm$ std across 5 seeds. Swimmer at 500K steps; all others at 1M steps.}
\label{tab:patch-method}
\small
\begin{tabular}{@{}lccccc@{}}
\toprule
Method & Swimmer & Hopper & Walker2d & HalfCheetah & Ant \\
\midrule
MoE-GUIDE    & $342 \pm 2$   & $3540 \pm 86$   & $4146 \pm 877$  & $10646 \pm 657$ & $3974 \pm 1099$ \\
\midrule
Regression   & $339 \pm 3$   & $3544 \pm 76$   & $4535 \pm 722$  & $10097 \pm 457$ & $3827 \pm 681$  \\
DBSCAN       & $250 \pm 4$   & $3542 \pm 94$   & $4186 \pm 293$  & $9864 \pm 693$  & $3917 \pm 498$  \\
\bottomrule
\end{tabular}
\end{table}

\subsection{OOB Margin Ablation}
\label{sec:margin-ablation}

During FK kernel construction, the governance loop identifies
\emph{out-of-bounds} (OOB) observations---those whose observation
dimensions fall outside the demonstrated $[\min, \max]$ range---and
places suppression patches at the corresponding bottleneck embeddings.
The \textbf{margin} parameter $m$ relaxes this detection boundary by
$m$ standard deviations per dimension:
\begin{equation}
\text{effective}_\text{lo} = \min - m \cdot \sigma, \qquad
\text{effective}_\text{hi} = \max + m \cdot \sigma,
\end{equation}
where $\sigma$ is the per-dimension standard deviation of the demo data.
With $m = 0$ (no margin), any observation outside the exact demo range
triggers patching; with $m > 0$, observations must deviate further
before patches are placed, reducing overly aggressive suppression
near the demonstration boundary.

\begin{figure}[H]
\centering
\includegraphics[width=\linewidth]{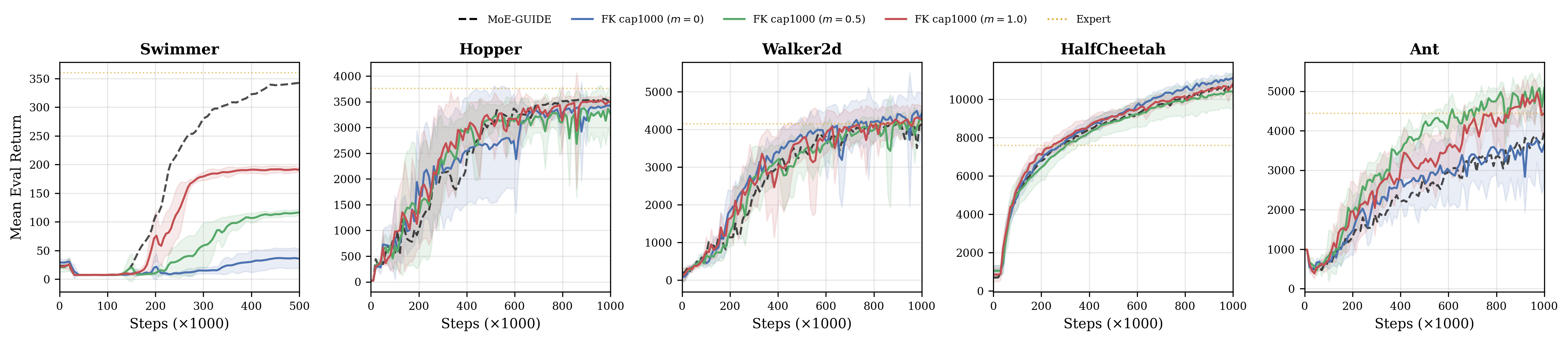}
\caption{OOB margin ablation across five MuJoCo environments.
Bars show mean evaluation return ($\pm$ std) for FK kernels with margin
$m \in \{0, 0.5, 1.0\}$.
Dashed line: MoE-GUIDE baseline; dotted line: expert demonstration return.}
\label{fig:margin-ablation}
\end{figure}

Table~\ref{tab:margin-ablation} compares FK kernels built with margin
$m \in \{0, 0.5, 1.0\}$, all using the regression patch-identification
method and a 1{,}000-patch cap.
Each environment uses its best tightness level
(Swimmer: very loose; all others: tight).
Figure~\ref{fig:margin-ablation} visualises the results.

\begin{table}[t]
\centering
\caption{OOB margin ablation: final mean evaluation return $\pm$ std
across 5 seeds.
Swimmer at 500K steps; all others at 1M steps.
All variants use regression patch identification with cap 1{,}000.
Each environment uses its best tightness level.
Bold indicates best FK variant per environment.}
\label{tab:margin-ablation}
\small
\begin{tabular}{@{}lccccc@{}}
\toprule
Method & Swimmer & Hopper & Walker2d$^\ddagger$ & HalfCheetah$^\ddagger$ & Ant$^\ddagger$ \\
\midrule
MoE-GUIDE         & $342 \pm 2$ & $3540 \pm 86$ & $4146 \pm 877$ & $10646 \pm 657$ & $3974 \pm 1099$ \\
\midrule
FK ($m{=}0.0$)     & $36 \pm 17$ & $3423 \pm 101$ & $\mathbf{4424 \pm 536}$ & $\mathbf{11096 \pm 216}$ & $3647 \pm 926$ \\
FK ($m{=}0.5$)     & $117 \pm 5$ & $3265 \pm 536$ & $4260 \pm 284$ & $10382 \pm 819$ & $\mathbf{5086 \pm 217}$ \\
FK ($m{=}1.0$)     & $\mathbf{191 \pm 4}$ & $\mathbf{3495 \pm 116}$ & $4268 \pm 344$ & $10800 \pm 524$ & $4443 \pm 370$ \\
\bottomrule
\end{tabular}
\end{table}

\begin{figure}[H]
\centering
\includegraphics[width=\linewidth]{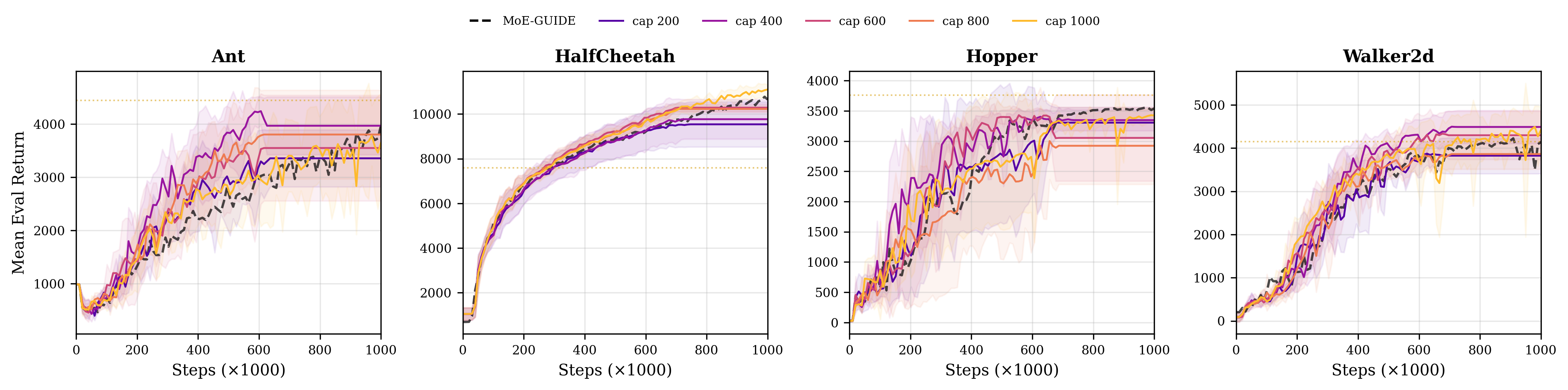}
\caption{Effect of OOB margin on FK Patching in Swimmer (left) and Ant (right).
Swimmer shows progressive recovery from collapse as margin increases.
Ant peaks at $m{=}0.5$, achieving the highest FK result across all ablations.}
\label{fig:margin-focus}
\end{figure}

The margin has a pronounced effect in two environments
(Figure~\ref{fig:margin-focus}):

\begin{itemize}
  \item \textbf{Swimmer} ($d_\text{obs} = 8$):
    Without margin, FK with cap~1{,}000 collapses to $36 \pm 17$---well
    below both MoE-GUIDE ($342$) and SAC ($78$).
    Zero-margin patching is overly aggressive in this low-dimensional,
    tight-distribution environment: too many near-boundary observations
    receive suppression, starving the agent of intrinsic reward.
    Increasing the margin progressively recovers performance
    ($m{=}0.5$: $117$; $m{=}1.0$: $191$), though neither
    fully reaches the MoE-GUIDE baseline.
    Note that the uncapped FK variant used in
    Table~\ref{tab:main-results} ($341 \pm 3$) does not suffer this
    collapse, indicating that the interaction between the patch cap
    and margin determines which observations receive patches.

  \item \textbf{Ant} ($d_\text{obs} = 111$):
    Margin $m{=}0.5$ achieves $5086 \pm 217$, the highest FK result
    across all ablations---surpassing both the no-margin cap-1{,}000
    variant ($3647 \pm 926$, $+39\%$) and the uncapped FK Patching
    entry in Table~\ref{tab:main-results} ($4816 \pm 594$).
    A moderate relaxation of the OOB boundary in the high-dimensional
    Ant observation space prevents patch overgeneralisation while still
    suppressing genuinely out-of-distribution states.
\end{itemize}

In Hopper, Walker2d, and HalfCheetah the differences are within one
standard deviation, indicating that the margin is less critical in
these environments.
HalfCheetah favours $m{=}0$, consistent with its strong performance
under the original FK kernel.

\begin{figure}[H]
\centering
\includegraphics[width=\linewidth]{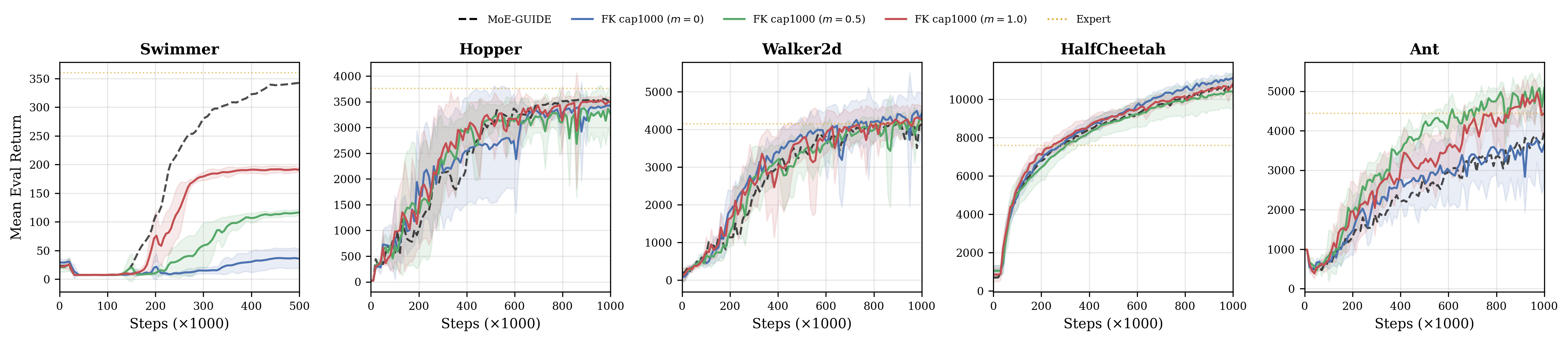}
\caption{Tightness--margin interaction for Ant-v2 (left) and HalfCheetah-v2 (right).
Ant benefits from margin across all tightness levels.
HalfCheetah performs best without margin at tight tightness.}
\label{fig:margintightness}
\end{figure}

Table~\ref{tab:margintightness} and Figure~\ref{fig:margintightness}
report additional tightness--margin combinations for Ant and HalfCheetah,
which had the most kernel variants available.
In Ant, the margin consistently improves over the corresponding
no-margin baseline across all tightness levels.

\begin{table}[t]
\centering
\label{tab:margintightness}

\small

\begin{tabular}{@{}llcc@{}}
\toprule
Tightness & Margin & Ant$^\ddagger$ & HalfCheetah$^\ddagger$ \\
\midrule
Tight  & $m{=}0.0$   & $3647 \pm 926$ & $\mathbf{11096 \pm 216}$ \\
Tight  & $m{=}0.5$   & $\mathbf{5086 \pm 217}$ & $10382 \pm 819$ \\
Tight  & $m{=}1.0$   & $4443 \pm 370$ & $10800 \pm 524$ \\
\midrule
Medium  & $m{=}1.0$   & $4680 \pm 835$ & $10542 \pm 612$ \\
Loose  & $m{=}1.0$   & $4263 \pm 946$ & $10790 \pm 606$ \\
\bottomrule
\end{tabular}
\end{table}

\section{Bottleneck Locality Ablations}
\label{app:bottleneck_locality}

This section provides the full per-coverage-level results for the bottleneck-versus-raw-space ablation. The experiment measures how many greedy ball-cover centers are needed to cover 2{,}000 OOB test points at several coverage thresholds. The ball radius is the median 10-nearest-neighbor distance in each respective space. Three conditions are compared: bottleneck space with its own radius (BN), raw observation space with its own radius (Raw), and raw space with the bottleneck radius (Raw@BN-$r$).

\begin{table}[h]
\centering
\caption{Ant ($D=111\to B=10$, 3 experts): patches at each coverage level.}
\label{tab:bn_raw_ant}
\small
\begin{tabular}{rrrrrr}
\toprule
Coverage & BN & Raw & Raw@BN-$r$ & Ratio & CoD \\
\midrule
50\% &  37 &  26 &  996 & 0.7$\times$ & 26.9$\times$ \\
75\% & 192 & 322 & 1496 & 1.7$\times$ &  7.8$\times$ \\
90\% & 492 & 622 & 1796 & 1.3$\times$ &  3.7$\times$ \\
95\% & 592 & 722 & 1896 & 1.2$\times$ &  3.2$\times$ \\
\bottomrule
\end{tabular}
\end{table}

\begin{table}[h]
\centering
\caption{HalfCheetah ($D=17\to B=7$, 4 experts): patches at each coverage level.}
\label{tab:bn_raw_halfcheetah}
\small
\begin{tabular}{rrrrrr}
\toprule
Coverage & BN & Raw & Raw@BN-$r$ & Ratio & CoD \\
\midrule
50\% &  35 &  30 &  885 & 0.9$\times$ & 25.3$\times$ \\
75\% & 128 & 160 & 1385 & 1.2$\times$ & 10.8$\times$ \\
90\% & 340 & 436 & 1685 & 1.3$\times$ &  5.0$\times$ \\
95\% & 440 & 536 & 1785 & 1.2$\times$ &  4.1$\times$ \\
\bottomrule
\end{tabular}
\end{table}

\begin{table}[h]
\centering
\caption{Hopper ($D=11\to B=4$, 2 experts): patches at each coverage level.}
\label{tab:bn_raw_hopper}
\small
\begin{tabular}{rrrrrr}
\toprule
Coverage & BN & Raw & Raw@BN-$r$ & Ratio & CoD \\
\midrule
50\% &  47 &  39 &  988 & 0.8$\times$ & 21.0$\times$ \\
75\% & 129 & 157 & 1488 & 1.2$\times$ & 11.5$\times$ \\
90\% & 284 & 402 & 1788 & 1.4$\times$ &  6.3$\times$ \\
95\% & 384 & 502 & 1888 & 1.3$\times$ &  4.9$\times$ \\
\bottomrule
\end{tabular}
\end{table}

\begin{table}[h]
\centering
\caption{Swimmer ($D=8\to B=3$, 1 expert): patches at each coverage level.}
\label{tab:bn_raw_swimmer}
\small
\begin{tabular}{rrrrrr}
\toprule
Coverage & BN & Raw & Raw@BN-$r$ & Ratio & CoD \\
\midrule
50\% &  57 &  44 &  967 & 0.8$\times$ & 17.0$\times$ \\
75\% & 127 & 156 & 1467 & 1.2$\times$ & 11.6$\times$ \\
90\% & 227 & 397 & 1767 & 1.7$\times$ &  7.8$\times$ \\
95\% & 313 & 497 & 1867 & 1.6$\times$ &  6.0$\times$ \\
\bottomrule
\end{tabular}
\end{table}

\begin{table}[h]
\centering
\caption{Walker2d ($D=17\to B=7$, 4 experts): patches at each coverage level.}
\label{tab:bn_raw_walker}
\small
\begin{tabular}{rrrrrr}
\toprule
Coverage & BN & Raw & Raw@BN-$r$ & Ratio & CoD \\
\midrule
50\% &  48 &  34 &  981 & 0.7$\times$ & 20.4$\times$ \\
75\% & 150 & 188 & 1481 & 1.3$\times$ &  9.9$\times$ \\
90\% & 385 & 488 & 1781 & 1.3$\times$ &  4.6$\times$ \\
95\% & 485 & 588 & 1881 & 1.2$\times$ &  3.9$\times$ \\
\bottomrule
\end{tabular}
\end{table}

\begin{table}[h]
\centering
\caption{AntMaze ($D=29\to B=8$, 3 experts): patches at each coverage level.}
\label{tab:bn_raw_antmaze}
\small
\begin{tabular}{rrrrrr}
\toprule
Coverage & BN & Raw & Raw@BN-$r$ & Ratio & CoD \\
\midrule
50\% &  39 &  29 &  998 & 0.7$\times$ & 25.6$\times$ \\
75\% & 159 & 184 & 1498 & 1.2$\times$ &  9.4$\times$ \\
90\% & 408 & 484 & 1798 & 1.2$\times$ &  4.4$\times$ \\
95\% & 508 & 584 & 1898 & 1.1$\times$ &  3.7$\times$ \\
\bottomrule
\end{tabular}
\end{table}

\begin{table}[h!]
\centering
\caption{Average neighbours within the ball radius.}
\label{tab:neighbourhood_stats}
\small
\begin{tabular}{lrrrrrr}
\toprule
Env & $D$ & $B$ & BN radius & Raw radius & Avg neigh (BN) & Avg neigh (Raw@BN-$r$) \\
\midrule
Ant         & 111 & 10 & 1.22 & 4.60 & 21.1 & 1.0 \\
HalfCheetah &  17 &  7 & 0.96 & 5.35 & 23.3 & 1.2 \\
Hopper      &  11 &  4 & 0.45 & 3.91 & 15.4 & 1.0 \\
Swimmer     &   8 &  3 & 0.25 & 2.56 & 11.8 & 1.0 \\
Walker2d    &  17 &  7 & 1.17 & 5.15 & 16.0 & 1.0 \\
AntMaze     &  29 &  8 & 1.15 & 7.55 & 29.2 & 1.0 \\
\bottomrule
\end{tabular}
\end{table}

At 50\% coverage, bottleneck space can appear worse than raw space because the first few greedy centers in raw space cover more points when the raw radius is large. At 75\% coverage and above, the bottleneck consistently requires fewer patches. The Raw@BN-$r$ condition shows the cost of using a fine bottleneck-scale radius in raw space: nearly every test point needs its own patch because fixed-radius balls become sparse in high-dimensional observation space.

\subsection{Bottleneck Dimension Sweep}
\label{app:patching:bn_dim}

This ablation varies the bottleneck dimension $B$ for Ant ($D=111$) while holding the number of experts fixed at 3. For each $B$, a fresh MoE is trained for 2{,}000 epochs with Adam (lr $=10^{-3}$), MSE loss, and full-batch training on all 26{,}594 demonstration observations. Patch efficiency is measured by greedy ball-cover over 2{,}000 OOB test points with median 10-NN radius.

\begin{table}[h!]
\centering
\caption{Full bottleneck dimension sweep for Ant ($D=111$, 3 experts, 2{,}000 epochs, 2{,}000 OOB test points).}
\label{tab:bn_dim_full}
\small
\begin{tabular}{rrrrrrrrrr}
\toprule
$B$ & Loss & Avg neigh & $P_{50}$ & $P_{75}$ & $P_{90}$ & $P_{95}$ & Final $P$ & Final cov. & Train (s) \\
\midrule
  2 & 0.0999 & 13.5 &  51 & 125 & 241 & 336 &  416 & 99\% & 124 \\
  3 & 0.0430 & 17.7 &  47 & 136 & 311 & 411 &  491 & 99\% & 127 \\
  5 & 0.0080 & 19.5 &  42 & 159 & 408 & 508 &  588 & 99\% & 126 \\
  7 & 0.0032 & 29.1 &  35 & 188 & 487 & 587 &  667 & 99\% & 139 \\
 10 & 0.0026 & 25.5 &  37 & 191 & 489 & 589 &  669 & 99\% & 136 \\
 15 & 0.0017 & 26.4 &  37 & 242 & 542 & 642 &  722 & 99\% & 152 \\
 20 & 0.0014 & 24.6 &  33 & 237 & 537 & 637 &  717 & 99\% & 148 \\
 30 & 0.0008 & 31.4 &  33 & 239 & 539 & 639 &  719 & 99\% & 189 \\
 50 & 0.0002 & 25.5 &  34 & 263 & 563 & 663 &  743 & 99\% & 221 \\
 80 & 0.0003 & 28.1 &  31 & 247 & 547 & 647 &  727 & 99\% & 271 \\
111 & ---    & 42.5 &  26 & 322 & 622 & 722 &  802 & 99\% & --- \\
\bottomrule
\end{tabular}
\end{table}

The sweep shows three regimes. Aggressive compression ($B\leq5$) maximizes patch efficiency but has higher reconstruction loss. The range $B\in[7,10]$ gives low reconstruction loss and stable patch counts. For $B>10$, additional dimensions add little reconstruction accuracy while increasing the intrinsic dimensionality of the patch space.

\section{Constraint-Targeted Patching and OOB Enforcement}
\label{app:constraint_results}

This section provides the complete Ant-v2 constraint analysis summarized in the main paper. Results are reproduced by \texttt{reproduce\_constraint\_analysis.py} from cached data in \texttt{ant\_constraint\_cache.pkl}.

\subsection{Experimental Setup}
\label{app:constraint-setup}

We define three Ant-v2 constraints inspired by Safety Gymnasium: speed $\sqrt{v_x^2+v_y^2}\leq6.0$, torso inclination from upright $\leq45^\circ$, and hip velocity $\max_i |v_{\text{hip},i}|\leq5.0$. Constraint violations are measured per evaluation episode as the fraction of transitions violating each constraint. The any-rate is the fraction of transitions violating at least one constraint. Constraint episodes are paired with the nearest evaluation checkpoint within $\pm20$K steps.

The demonstration dataset used for this constraint analysis contains 26{,}594 observations from 12 trajectory files. Of these, 85.8\% violate at least one constraint, reflecting the nature of Ant-v2 locomotion: agile movement involves high speeds, tilting, and rapid joint motion.

We evaluate Constraint patching (CP), Flywheel patching (FP), and OOB regimes. CP places local suppression patches in MoE bottleneck space at constraint-violating demonstration observations. Flywheel patching uses high-reconstruction-error flywheel patches across all observation dimensions with a cap of 1000 patches. OOB checks all 27 active observation dimensions against the demonstrated range, with hard clipping or smooth Gaussian decay, and optional safety functions.

\subsection{Violation Rates by Training Phase}
\label{app:violation-phase}

Table~\ref{tab:violation-phase} shows how violation rates evolve during training. CP methods start with violation rates comparable to MoE-GUIDE and show no consistent decrease over time. OOB methods show an initial dip followed by a gradual increase as the policy learns higher-return behavior.

\begin{table}[H]
\centering
\caption{Violation rate (any\_rate) by training phase.}
\label{tab:violation-phase}
\small
\begin{tabular}{@{}lrrrr@{}}
\toprule
Method & 0--100K & 100--200K & 200--400K & 400--600K \\
\midrule
SAC               & 0.804 & 0.806 & 0.808 & 0.801 \\
MoE-GUIDE         & 0.734 & 0.690 & 0.776 & 0.794 \\
Fp tight          & 0.734 & 0.645 & 0.655 & 0.684 \\
\midrule
CP: Combined      & 0.785 & 0.773 & 0.785 & 0.822 \\
CP: Velocity      & 0.770 & 0.763 & 0.806 & 0.820 \\
CP: Tilt          & 0.792 & 0.780 & 0.813 & 0.839 \\
CP: Hip Vel       & 0.791 & 0.640 & 0.748 & 0.799 \\
\midrule
OOB Hard+Comb.    & 0.485 & 0.095 & 0.170 & 0.290 \\
OOB Hard+Tilt     & 0.512 & 0.173 & 0.415 & 0.604 \\
OOB Smooth+Comb.  & 0.492 & 0.160 & 0.178 & 0.244 \\
OOB Smooth+Tilt   & 0.561 & 0.271 & 0.464 & 0.617 \\
\bottomrule
\end{tabular}
\end{table}

\subsection{Per-Constraint Violation Rates}
\label{app:per-constraint}

\begin{table}[H]
\centering
\caption{Per-constraint violation rates, global average across all steps and seeds.}
\label{tab:per-constraint}
\small
\begin{tabular}{@{}lrrr@{}}
\toprule
Method & Speed & Tilt & Hip vel \\
\midrule
SAC               & 0.001 & 0.430 & 0.534 \\
MoE-GUIDE         & 0.027 & 0.473 & 0.469 \\
Fp tight          & 0.014 & 0.375 & 0.460 \\
\midrule
CP: Combined      & 0.033 & 0.457 & 0.529 \\
CP: Velocity      & 0.019 & 0.479 & 0.503 \\
CP: Tilt          & 0.027 & 0.519 & 0.503 \\
CP: Hip Vel       & 0.001 & 0.466 & 0.399 \\
\midrule
OOB Hard+Comb.    & 0.014 & 0.148 & 0.262 \\
OOB Hard+Tilt     & 0.008 & 0.136 & 0.446 \\
OOB Smooth+Comb.  & 0.016 & 0.128 & 0.273 \\
OOB Smooth+Tilt   & 0.024 & 0.166 & 0.452 \\
\bottomrule
\end{tabular}
\end{table}

\subsection{Reward--Violation Correlation}
\label{app:correlation}

Speed violations show a strong positive correlation with reward across methods, confirming that faster locomotion yields higher returns and is difficult to suppress without affecting performance. Tilt shows weaker and sometimes negative correlations, suggesting it can be reduced more independently.

\begin{table}[H]
\centering
\caption{Pearson $r$ between evaluation reward and violation rate.}
\label{tab:correlation}
\small
\begin{tabular}{@{}lrrrr@{}}
\toprule
Method & $r$(any) & $r$(speed) & $r$(tilt) & $r$(hip) \\
\midrule
SAC               & $-0.02$ & $+0.33$ & $-0.27$ & $+0.31$ \\
MoE-GUIDE         & $+0.25$ & $+0.70$ & $+0.14$ & $+0.25$ \\
FP tight          & $+0.08$ & $+0.66$ & $+0.02$ & $+0.20$ \\
\midrule
CP: Combined      & $+0.20$ & $+0.74$ & $+0.17$ & $+0.16$ \\
CP: Tilt          & $+0.16$ & $+0.68$ & $+0.10$ & $+0.25$ \\
\midrule
OOB Hard+Comb.    & $+0.54$ & $+0.81$ & $+0.22$ & $+0.52$ \\
OOB Smooth+Comb.  & $+0.50$ & $+0.75$ & $-0.00$ & $+0.52$ \\
\bottomrule
\end{tabular}
\end{table}

\subsection{Patch Suppression Mechanism Verification}
\label{app:patch-suppression}

To verify that CP patches function as designed, we evaluate patched and unpatched MoE models directly on the demonstration dataset, split into constraint-satisfying observations ($n=3{,}784$) and constraint-violating observations ($n=22{,}810$). Table~\ref{tab:patch-suppression} confirms that patches suppress violated observations while leaving safe observations virtually unchanged, but coverage is limited: CP Combined achieves meaningful suppression ($s>0.1$) on 21.4\% of violated demo observations.

\begin{table}[H]
\centering
\caption{Patch suppression on demonstration data. $r_\text{base}$ is unpatched MoE reward, $s$ is suppression magnitude, $r_\text{final}$ is reward after suppression, and $\Delta r$ is change from unpatched.}
\label{tab:patch-suppression}
\small
\begin{tabular}{@{}lr rrrrr rr@{}}
\toprule
& & \multicolumn{5}{c}{Violated ($n = 22{,}810$)} & \multicolumn{2}{c}{Safe ($n = 3{,}784$)} \\
\cmidrule(lr){3-7} \cmidrule(lr){8-9}
Kernel & Patches & $r_\text{base}$ & $s$ & $r_\text{final}$ & $\Delta r$ & $\%{>}0.1$ & $s$ & $\Delta r$ \\
\midrule
CP Combined  & 1{,}063 & 0.633 & 0.137 & 0.539 & $-0.094$ & 21.4\% & 0.000 & $-0.000$ \\
CP Velocity  &     416 & 0.633 & 0.011 & 0.624 & $-0.009$ &  3.9\% & 0.000 & $-0.000$ \\
CP Tilt      &     735 & 0.633 & 0.027 & 0.608 & $-0.025$ &  7.2\% & 0.000 & $-0.000$ \\
CP Hip Vel   &     699 & 0.633 & 0.121 & 0.557 & $-0.076$ & 15.4\% & 0.000 & $-0.000$ \\
FP cap1000   &     922 & 0.633 & 0.000 & 0.633 & $-0.000$ &  0.0\% & 0.000 & $-0.000$ \\
\bottomrule
\end{tabular}
\end{table}

Two factors explain why CP patches reduce violations in IL-only evaluation but not during RL training: the patch budget covers only a fraction of the violation manifold, and environment reward dominates the intrinsic suppression scale.

\subsection{Matched Reward Levels and OOB Variants}
\label{app:matched_reward_oob}

\begin{table}[H]
\centering
\caption{Violation rate (any\_rate) at matched reward levels.}
\label{tab:matched-violations}
\small
\begin{tabular}{@{}lrrrr@{}}
\toprule
Method & $R \in [500, 1K)$ & $R \in [1K, 1.5K)$ & $R \in [2K, 2.5K)$ & $R \in [2.5K, 3.5K)$ \\
\midrule
SAC               & 0.811 & 0.811 & 0.803 & 0.791 \\
MoE-GUIDE         & 0.687 & 0.748 & 0.788 & 0.809 \\
FP tight          & 0.698 & 0.614 & 0.607 & 0.667 \\
\midrule
CP: Combined      & 0.783 & 0.773 & 0.779 & 0.769 \\
CP: Tilt          & 0.796 & 0.771 & 0.805 & 0.835 \\
\midrule
OOB Hard+Comb.    & 0.194 & 0.312 & 0.586 & 0.638 \\
OOB Smooth+Comb.  & 0.218 & 0.213 & 0.482 & 0.630 \\
\bottomrule
\end{tabular}
\end{table}

\begin{table}[H]
\centering
\caption{Full OOB results: final performance over the last 100K steps.}
\label{tab:oob-full}
\small
\begin{tabular}{@{}lrrr@{}}
\toprule
Method & Steps $R{\geq}2K$ & Final $R$ & Final Viol. \\
\midrule
OOB Hard (no safety)    & 354K\,(5) & 2379 & --- \\
OOB Hard+Combined       & 447K\,(3) & 3590 & 0.677 \\
OOB Hard+Velocity       & 308K\,(5) & 3675 & 0.773 \\
OOB Hard+Tilt           & 386K\,(5) & 3253 & 0.655 \\
OOB Hard+Hip Vel        & 590K\,(5) & 2860 & 0.820 \\
\midrule
OOB Smooth (no safety)  & 316K\,(5) & 2094 & --- \\
OOB Smooth+Combined     & 578K\,(5) & 3332 & 0.577 \\
OOB Smooth+Velocity     & 460K\,(5) & 3579 & 0.803 \\
OOB Smooth+Tilt         & 466K\,(5) & 3872 & 0.654 \\
OOB Smooth+Hip Vel      & 435K\,(4) & 2684 & 0.812 \\
\bottomrule
\end{tabular}
\end{table}

\section{Slow Verifier Motivation}
\label{app:latency}

Tables~\ref{tab:vision_latency} and~\ref{tab:safety_latency} report representative latency measurements. They motivate offline compilation but are not part of an implemented image-based system in this paper.

\begin{table}[H]
\centering
\caption{Per-frame latency: vision pipelines vs. patch evaluation (single CPU core, $640\times480$ image, median-of-5).}
\label{tab:vision_latency}
\small
\begin{tabular}{lrrl}
\toprule
Pipeline stage & Latency (ms) & vs. Patches & Realtime? \\
\midrule
\textbf{MoE + 800 patches (brute)} & \textbf{0.36} & $1\times$ & $\checkmark$~all freqs \\
\textbf{MoE + 800 patches (LSH)} & \textbf{0.40} & $1.1\times$ & $\checkmark$~all freqs \\
\addlinespace
MobileNetV2 classify & 19.8 & $55\times$ & $\leq$50\,Hz only \\
ResNet-50 classify & 62.4 & $173\times$ & $\leq$15\,Hz only \\
KeypointRCNN (R50-FPN) & 5{,}772 & $16{,}033\times$ & $<$1\,Hz \\
Classify + Keypoint & 5{,}864 & $16{,}289\times$ & $<$1\,Hz \\
\bottomrule
\end{tabular}
\end{table}

\begin{table}[H]
\centering
\caption{Safety-signal latency spectrum: from formal logic to human review. Patches compile offline signals into runtime patch lookup.}
\label{tab:safety_latency}
\small
\begin{tabular}{llrl}
\toprule
Safety signal & Example & Latency & Realtime? \\
\midrule
Formal logic  & Bound check, velocity cap & $<0.01$\,ms & $\checkmark$ \\
Light neural net & MobileNetV2 classify & ${\sim}20$\,ms & Marginal \\
Heavy neural net & Keypoint + detection & ${\sim}6{,}000$\,ms & $\times$ \\
Multi-model consensus & Ensemble + voting & ${\sim}10{,}000$\,ms & $\times$ \\
Simulation rollout & Forward dynamics check & ${\sim}30{,}000$\,ms & $\times$ \\
Human review & Norm / ethical judgment & minutes--hours & $\times$ \\
\addlinespace
\textbf{Compiled patch} & \textbf{Any of the above} & $\mathbf{<0.5}$\,\textbf{ms} & $\checkmark$ \\
\bottomrule
\end{tabular}
\end{table}

\section{Implementation and artifact Format}
\label{app:implementation}

Each kernel is serialized as a PyTorch \texttt{.pt} file via \texttt{torch.save}. Table~\ref{tab:kernel_format} lists the stored fields. The kernel includes model weights, model configuration, reward-mapping parameters, dimension statistics, patch records, strictness metadata, environment name, and number of demonstration points.

\begin{table}[H]
\centering
\caption{Kernel file contents.}
\label{tab:kernel_format}
\small
\begin{tabular}{ll}
\toprule
Key & Description \\
\midrule
\texttt{moe\_state\_dict}  & MoE model weights \\
\texttt{moe\_config}       & \texttt{\{input\_dim, bottleneck\_dim, num\_experts\}} \\
\texttt{estimator\_config} & \texttt{\{l\_min, l\_max, steepness\}} \\
\texttt{dim\_stats}        & Per-dimension \texttt{\{min, max, mean, std\}} tensors \\
\texttt{dim\_groups}       & Position and velocity dimension groups \\
\texttt{flywheel\_patches} & Per-expert patch centers, bandwidths, and strengths \\
\texttt{strictness}        & Strictness label and associated patch parameters \\
\texttt{env\_name}         & Environment name string \\
\texttt{n\_demo\_points}   & Number of demonstration observations used \\
\bottomrule
\end{tabular}
\end{table}

The flywheel roles exchange typed artifacts. \texttt{CandidateFlaw} carries the observation vector, OOB ratio, violated group, safety value, uncertainty estimate, and oracle version. \texttt{VerificationResult} records the verification outcome, violated norm ID, and evidence string. \texttt{GovernanceBatch} contains local corrections, regression evidence, oracle-version transitions, and deployment signatures. These records provide an audit trail from candidate discovery to patch deployment.

\newpage
\bibliography{main}
\bibliographystyle{splncs04}

\end{document}